%% file: iclr2027_conference.tex
\documentclass{article} 
\usepackage{iclr2027_conference,times}

\iclrfinalcopy
\input{math_commands.tex}

\usepackage{hyperref}
\usepackage{url}
\usepackage{fontawesome5} 
\usepackage{graphicx}
\usepackage{xspace}
\usepackage{booktabs}
\usepackage[table]{xcolor}
\usepackage[most]{tcolorbox}
\usepackage{enumitem}
\usepackage{multirow}
\usepackage{longtable,array}
\usepackage{wrapfig}
\definecolor{promptboxbg}{RGB}{248,248,248}

\usepackage{float}

\usepackage{color-edits}
\addauthor{gn}{magenta}

\usepackage[most]{tcolorbox}
\definecolor{suggcol}{HTML}{3B8F8A}  
\definecolor{divcol}{HTML}{4F7FBF}   
\newtcbox{\suggbox}{on line, colback=suggcol!12, colframe=suggcol!45, coltext=suggcol!85!black,
  fontupper=\bfseries, boxrule=0.4pt, arc=2pt, boxsep=0pt, left=2pt, right=2pt, top=1pt, bottom=1pt}
\newtcbox{\divbox}{on line, colback=divcol!12, colframe=divcol!45, coltext=divcol!85!black,
  fontupper=\bfseries, boxrule=0.4pt, arc=2pt, boxsep=0pt, left=2pt, right=2pt, top=1pt, bottom=1pt}
\title{\method: Text-Conditioned Diversity for Image Generation}

\author{Simran Khanuja\textsuperscript{1} \qquad
Rohit Gandikota\textsuperscript{2} \qquad
Gaurav Parmar\textsuperscript{1} \qquad
Graham Neubig\textsuperscript{1} \\[0.5em]
\textsuperscript{1}Carnegie Mellon University \qquad \textsuperscript{2}Northeastern University \\[0.5em]
Correspondence: \href{mailto:skhanuja@andrew.cmu.edu}{\texttt{skhanuja@andrew.cmu.edu}} \qquad
\href{https://image-study.org/t-dig/}{\faGlobe\,\textbf{image-study.org/t-dig}}
}

\input{defs}

\begin{document}

\maketitle

\input{sections/0-abstract}

\input{sections/1-introduction}

\input{sections/2-framework}

\input{sections/3-training}

\input{sections/4-experiments}

\input{sections/5-results}
\input{sections/6-related}
\input{sections/7-conclusion}
\input{sections/8-limitations}



\bibliography{iclr2027_conference}
\bibliographystyle{iclr2027_conference}

\input{sections/9-appendix}

\end{document}

%% file: math_commands.tex
\usepackage{amsmath,amsfonts,bm}

\def\eqref#1{equation~\ref{#1}}

\def\1{\bm{1}}

\DeclareMathAlphabet{\mathsfit}{\encodingdefault}{\sfdefault}{m}{sl}
\SetMathAlphabet{\mathsfit}{bold}{\encodingdefault}{\sfdefault}{bx}{n}



%% file: defs.tex
\definecolor{coralpink}{rgb}{0.97, 0.51, 0.47}
\definecolor{babyblueeyes}{rgb}{0.63, 0.79, 0.95}

\newcommand{\method}{\textsc{T-DIG}\xspace}



%% file: sections/0-abstract.tex
\begin{abstract}
Text-to-image (T2I) platforms typically return several images per prompt so users can explore different visual interpretations and choose one that matches their intent. However, T2I models are usually optimized for single-image quality which comes at a cost of multi-image diversity, limiting this exploration. Diversity is also contextual: one user wants variations in background, while other in artistic style, and the same user may later want variations in clothing. The generated images should differ along whatever axis the user cares about.
In this paper, we introduce \method, a \emph{text-conditioned} diversity framework for image generation and editing. \method is an iterative framework comprising multiple modules that lets users type prompts, inspect default generations, request for variations using natural language, and continue refining outputs until they're satisfied. At its core, \method uses an LLM policy that produces $K$ axis-conditioned prompts or edit instructions for a frozen generator or editor. We train this policy with GRPO using a text-conditioned diversity reward and alignment penalties. Because the image generator remain frozen, \method applies to open-weight, distilled, and proprietary models. We also release the \emph{first dataset} for text-conditioned prompt diversity in image generation, with the training and validation splits built from human-written axes and a test set drawn from real user interactions with \method. \method far exceeds baselines in improving text-conditioned diversity, while preserving base-prompt alignment across a variety of metrics and human judgements. We show that user-directed diversity exploration can be added to a wide range of image generation models, open and closed, without the need to access their weights.\footnote{Website: \url{https://image-study.org/t-dig/}}

\end{abstract}

%% file: sections/1-introduction.tex
\section{Introduction}
\label{sec:introduction}

Text-to-image platforms~\citep{midjourney,firefly} typically show four to eight images per prompt, so that the user has multiple options to choose from. However, different users may often want to see different kinds of variation, which T2I models don't typically support.
We conducted a human study where users were shown images for \emph{``people sitting at a table with laptops and books''}; one user asked for variation in ``laptops and camera angle'', another wanted to see ``people doing different things'', and a third wanted to vary ``people's ethnicity''. Requests also ranged from being highly specific, to fairly abstract and open-ended. For \emph{a dog playing in a park}, a user asked for different ``dog breeds'', whereas for \emph{a comic-book panel...}, they wanted ``crazy endings that would surprise them''.

\begin{figure}[t]
    \centering
    \includegraphics[width=\textwidth]{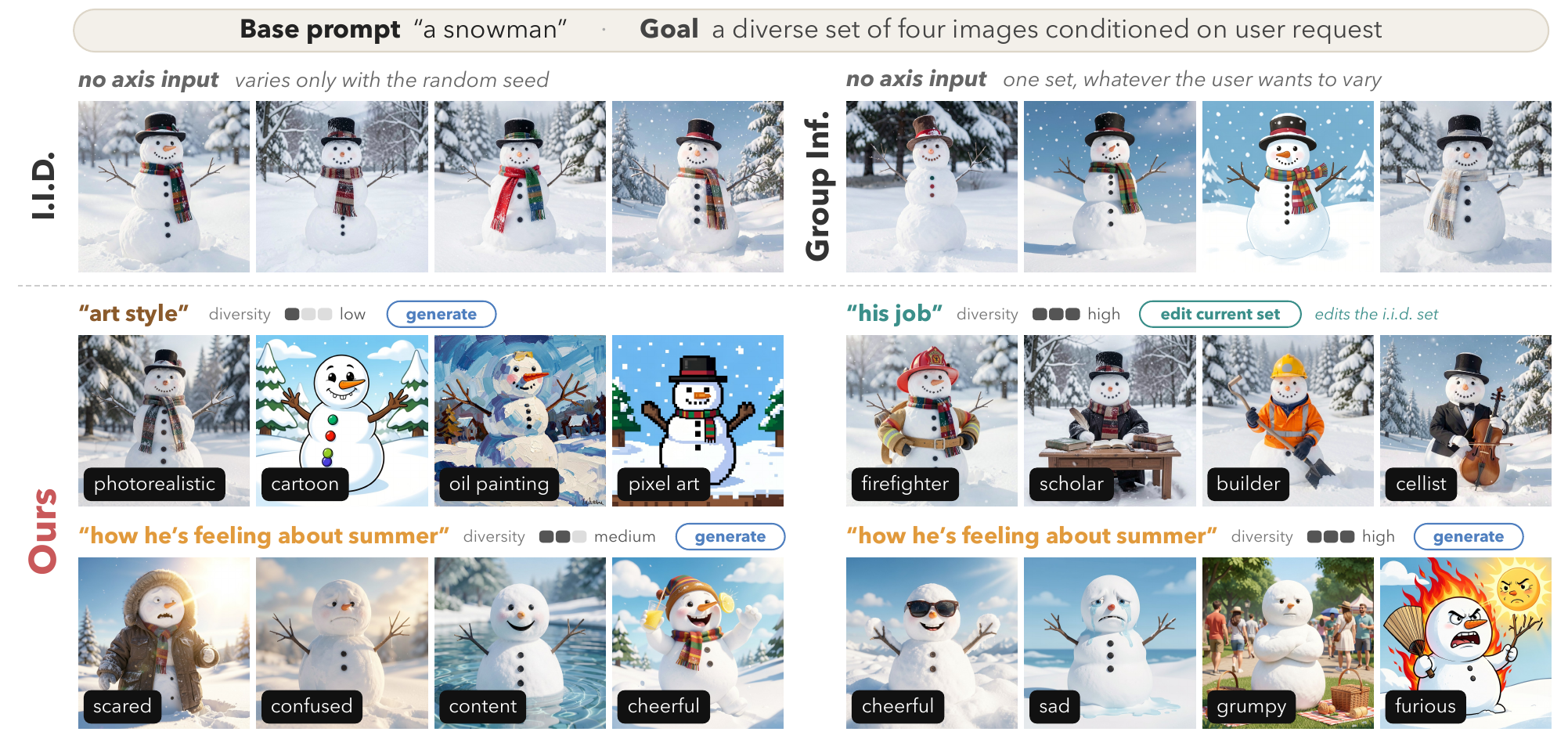}
    \caption{\textbf{Text-conditioned Diversity:}
    For one prompt, i.i.d. sampling and prior methods return one set regardless of what the user wants to vary; \method produces a different set for each text-condition, diversity strength, and can edit/generate, accounting for user preferences.}
    \label{fig:motivation}
\end{figure}

Prior works have proposed many methods to increase diversity of image generation models, but none can accommodate for these varying scenarios. One category of diversity methods~\citep{parmar2026scaling, corso2024particle} maximizes a pairwise distance in latent embedding spaces~\citep{fu2023dreamsim,oquab2023dinov2}. These approaches make the images generically diverse but do not take the user's request into consideration.
Some methods do target a specific proxy of diversity, for example, varying modes of an underspecified prompt~\citep{morshed2025diverseflow}, the identities of people in a scene~\citep{borse2026resolving}, or demographic and geographic coverage~\citep{askari2024improving,miao2024training}; but they do not generalize beyond that pre-specified target. All prior methods return one output set per prompt with a predetermined type of variation, as shown in Figure~\ref{fig:motivation}.

In this paper, we treat diversity as a conditional generation problem: given a prompt and an axis of variation written in free-form text, produce $K$ images that differ along that axis while following the prompt. We build \method, an interactive framework around this idea (Section~\ref{sec:framework}). At its core, an LLM policy expands the base prompt into $K$ axis-conditioned prompts or edit instructions for a frozen T2I model. We train the policy with GRPO, rewarding text-conditioned diversity measured with TPIPS~\citep{wang2026many} and penalizing loss of alignment with the base prompt. TPIPS is a VLM fine-tuned on human perceptual judgments to score image similarity along a free-form text-specified aspect, so it lets us measure diversity along user-specified axes. Because the T2I model stays frozen and need not be differentiable, \method applies to open-weight, distilled, and API-only generators and editors.

To the best of our knowledge, no existing dataset pairs prompts with user-written axes of variation. We therefore build training and validation data from HPDv2 prompts~\citep{wu2023human} with axes elicited from a VLM using human-written examples in context, and a test set of 405 (prompt, axis) pairs written by participants using \method. Across four open-weight T2I models, \method more than doubles diversity along the requested axis over i.i.d.\ sampling on average ($2.3\times$, and up to $3.1\times$), while lowering alignment with the base prompt by less than 2\%. Existing diversity methods (that do not condition on any axis), fall short on one side or the other: the most diverse baseline, CADS~\citep{sadat2024cads}, reaches $1.8\times$ at a 6\% alignment drop, while Scaling Group Inference~\citep{parmar2026scaling} preserves alignment best (under 1\% drop) but gains only $1.3\times$, at up to $22\times$ the sampling cost. Policies trained with open-weight models also transfer to proprietary ones -- on GPT Image 2, our Qwen-Image policy raises diversity about 40\% more than asking GPT-5.5 to produce varied prompts along the given axis. In pairwise human evaluation, raters find \method's sets more varied along the requested axis than every baseline's on 64--96\% of instances across all T2I models in our study. Our overall contributions are:


\begin{enumerate}
    \item \textbf{Task formulation:} We formulate T2I diversity as conditional generation, where users state in free-form text what should vary, from broad axes like style to prompt-specific requests no fixed taxonomy covers.

    \item \textbf{Interactive framework:} \method lets users request, control, and iteratively refine variations in natural language, by generating new sets or editing existing ones.

    \item \textbf{Model-agnostic training:} We train generation and editing policies with GRPO on a text-conditioned diversity reward and a baseline-relative alignment penalty; the T2I model stays frozen, so \method works with open-weight, distilled, and proprietary models.

    \item \textbf{User-derived dataset for text-conditioned diversity:} We release human-grounded training and validation data, along with a test set collected from real user interactions and a human rating study against three baselines.
\end{enumerate}

\vspace{-1mm}

%% file: sections/2-framework.tex
\begin{figure}[t]
    \centering
    \includegraphics[width=\textwidth]{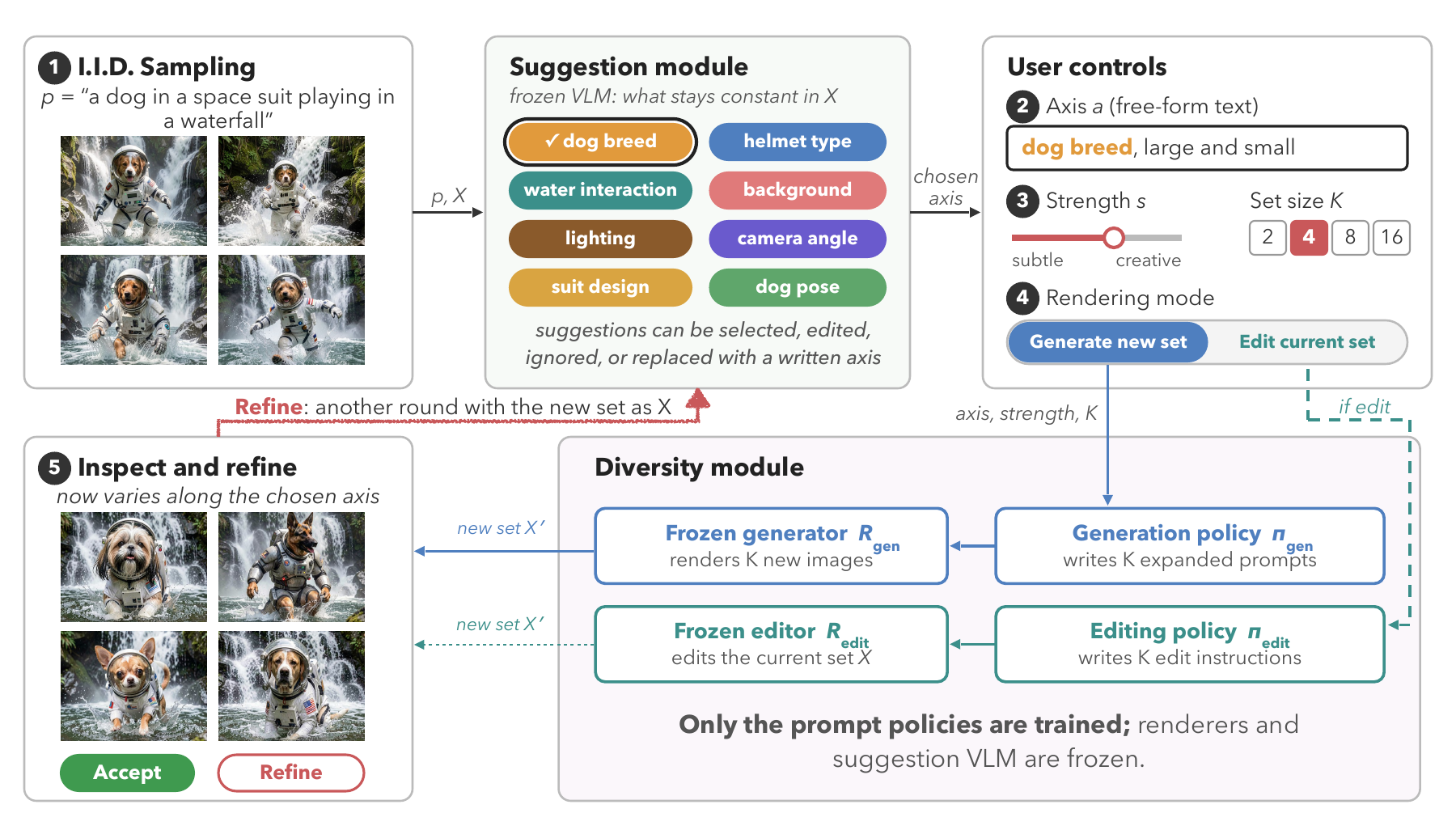}
    \caption{\textbf{\method framework.} After inspecting default samples, the user writes an axis or picks a suggested one, sets the diversity strength, and chooses to generate a new set or edit the current one. A trained policy writes $K$ prompts or edit instructions for a frozen T2I model or editor, and the user accepts the result or refines it for another round. Only the policies are trained.}
    \label{fig:framework}
\end{figure}

\section{Framework Design}
\label{sec:framework}
\vspace{-1mm}
Figure~\ref{fig:framework} shows how a user interacts with \method. We describe each step of interaction and the diversity module we train below:
\vspace{-2mm}
\paragraph{1. I.I.D Sampling:} The user enters a prompt $p$ and sees $K$ samples drawn with independent seeds. These show what the T2I model does by default; if the set is already diverse according to what the user wants, they stop here. Otherwise a frozen \suggbox{Suggestion Module}, given $p$ and the samples, suggests axes of variation based on the images generated and the input prompt.
\vspace{-2mm}
\paragraph{2. Axis:} The user writes the axis of variation $a$ in free-form text, or picks or edits a suggestion that we give. An axis can be broad (``artistic style'', ``composition''), specific (``the arrangement of people around the table''), or simply ``overall''.
\vspace{-2mm}

\paragraph{3. Strength and set size:} The user sets a set size $K\in\{2,4,8,16\}$ and a diversity strength $s$, which ranges from minimal substitutions along the axis to richer scenes that vary more.

\vspace{-2mm}
\paragraph{4. Rendering Mode:} The user can choose to either generate a new set of images that vary along the axis, or edit the one already generated if they want to add more variation to the current set.
\vspace{-2mm}
\paragraph{5. Inspect and refine:} The user accepts the new set $\mathcal{X}'$, or starts another round with $\mathcal{X}'$ as input, changing any of $a$, $s$, $K$, or the mode; and our suggestions are recomputed for the new set.

\divbox{Diversity Module} is a pair of post-trained LLM policies,
$\pi_{\text{gen}}$ and $\pi_{\text{edit}}$. Each is a LoRA adapter $\Delta W = BA$ on a frozen
language model with weights $W_0$, and strength $s$ sets its scale $\alpha_s \in [0,1]$,
giving the policy $\pi_{\alpha_s}$ with weights $W_0 + \alpha_s \Delta W$. Given $p$, $a$, and
$K$ (for editing additionally the current set $\mathcal{X}$), it writes $K$ prompts or edit
instructions:
\begin{equation}
  \mathbf{q} = \pi_{\alpha_s}(p, a, K; \mathcal{X}) = \{q_1, \ldots, q_K\},
  \label{eq:policy}
\end{equation}
which a frozen T2I model turns into the new set, $\mathcal{X}'=R_{\mathrm{gen}}(\mathbf{q})$ in generation mode and $\mathcal{X}'=R_{\mathrm{edit}}(\mathcal{X},\mathbf{q})$ in editing mode. Only the prompt-set adapters are trained; the generator, the editor, and the suggestion VLM stay frozen.

%% file: sections/3-training.tex
\section{Training}
\label{sec:training}
\vspace{-1mm}
In this section, we describe how we train the two LLM policies of the \divbox{Diversity Module} with GRPO, for generation and editing. 


\vspace{-1mm}
\subsection{Training Data}
\label{sec:training-data}

\vspace{-1mm}
\paragraph{Base prompts:}
We start from the HPDv2 training split~\citep{wu2023human}, which contains user-written T2I prompts. After removing empty entries and exact duplicates, 103,672 unique prompts remain. We filter this pool with GPT-5.6 Luna~\citep{openai2026gpt56luna} using medium reasoning effort. The filter removes prompts that depict named real individuals, contain unsafe content, lack a depictable subject, or are dominated by style and quality descriptors. A prompt is retained only if the model can identify at least three meaningful axes of variation. This step retains 79.9\% of the unique pool. Even after exact-duplicate removal and filtering, we find that many retained prompts are semantically similar. To reduce redundancy in training, we cluster the remaining prompts and choose a sample from each, leaving roughly 20,000 prompts (filtering and clustering details in Appendix~\ref{app:train-data}).

\vspace{-2mm}
\paragraph{Axes:} In a pilot study, participants saw i.i.d.\ outputs for 20 prompts and wrote what they would like to vary. This is used as seed data to elicit axes of variation from a VLM at scale. For each base prompt $p$ we collect above, we render $K$ i.i.d.\ samples with the T2I model and ask Gemini 3.8 Flash~\citep{google2026gemini38flashapi} for up to four free-form axes, using the pilot instruction and responses as in-context examples. For a random 10\% of prompts we also add the axis ``overall'', which supports axis-free requests in case people would like to see overall diversity. After removing malformed outputs and holding out 200 prompts for validation, the training set contains 19,765 prompts and 76,089 (prompt, axis) pairs.

\vspace{-1mm}
\subsection{Policy Training}
\label{sec:policy}

Both policies are trained with a set-level reward that favors diversity along the text condition while penalizing loss of alignment with the base prompt. We describe the different components that make up the final reward below:

\vspace{-2mm}
\paragraph{Text-conditioned diversity:}
For an image set $\mathcal{X}=\{x_1,\ldots,x_K\}$ and an axis in the form of text prompt $a$, TPIPS~\citep{wang2026many} embeds each image conditioned on the axis, $e_i=f_{\mathrm{TPIPS}}(x_i;a)$, which means TPIPS embeddings of images are similar when they look alike along $a$, even if they differ in other aspects. We measure the spread of these embeddings with the Vendi Score ($\mathrm{VS}$)~\citep{friedman2022vendi}; given unit-normalized TPIPS embeddings and their cosine-similarity matrix $S_{ij}=e_i^\top e_j$, $\mathrm{VS}=\exp\big(-\sum_k \lambda_k\log\lambda_k\big)$, where $\lambda_k$ are the eigenvalues of $S/K$. It essentially gives us the effective number of distinct images in the set, from 1 (all identical along $a$) to $K$ (all dissimilar). We normalize it by set size:
\begin{equation}
D_a(\mathcal{X}) = \frac{\mathrm{VS}(\{e_i\}_{i=1}^K) - 1}{K-1} \in [0,1].
\label{eq:diversity}
\end{equation}



\paragraph{Base-prompt alignment:}
We measure alignment with VQAScore~\citep{lin2024evaluating}. Alignment under i.i.d.\ sampling varies widely across prompts, so a single threshold would penalize many prompts before the policy changes anything. We therefore set a per-prompt threshold $\tau(p,R)$ to the mean VQAScore of $K$ i.i.d.\ samples from the renderer $R$ being trained for, and penalize only images that fall below their own i.i.d score threshold:
\begin{equation}
    \mathcal{L}_{\mathrm{align}}(\mathcal{X};p,R)=\frac{1}{K}\sum_{i=1}^{K}\max\big(0,\,\tau(p,R)-\mathrm{VQAScore}(x_i,p)\big).
\end{equation}

\vspace{-2mm}
\paragraph{Prompt length:} Without a length term, the policy learns to write overly detailed prompts that the alignment hinge does not penalize. This makes diversity leak off-axis, and text past the encoder's context limit is also truncated, hence a length penalty is needed. As with alignment, we set the budget relative to the base prompt where each expanded prompt may add at most $w$ words to $p$, $c(p)=|p|+w$, where $|\cdot|$ is the word count and $w$ is a hyperparameter:
\begin{equation}
    \mathcal{L}_{\mathrm{len}}(\mathbf{q};p)=\frac{1}{K}\sum_{i=1}^{K}\frac{\max\big(0,\,|q_i|-c(p)\big)}{c(p)}.
\end{equation}

\paragraph{Preference preservation:}
Even after alignment and length penalties, a policy can raise VQAScore while making the images worse, for example by asking the model to print the base prompt as a caption inside the image, which only human preference scores tend to penalize. We therefore add a hinge on HPSv3~\citep{ma2025hpsv3}, calibrated per prompt like the alignment threshold: $\kappa(p,R)$ is the mean HPSv3 of the $K$ i.i.d.\ samples $R(p,z_j)$, and
\begin{equation}
    \mathcal{L}_{\mathrm{pref}}(\mathcal{X};p,R)
    =
    \frac{1}{K}\sum_{i=1}^{K}\max\big(0,\,\kappa(p,R)-\operatorname{HPSv3}(x_i,p)\big).
\end{equation}

%


\paragraph{Final Reward:} The final reward combines the four terms above; $\lambda$, $\mu$ and $\eta$ trade diversity along text-condition ($a$), against alignment with base prompt ($p$), prompt length, and human preference:
\begin{equation}
    r(\mathbf{q},\mathcal{X};p,a,R)=D_a(\mathcal{X})-\lambda\,\mathcal{L}_{\mathrm{align}}(\mathcal{X};p,R)-\mu\,\mathcal{L}_{\mathrm{len}}(\mathbf{q};p)-\eta\,\mathcal{L}_{\mathrm{pref}}(\mathcal{X};p,R),
    \label{eq:reward}
\end{equation}

\paragraph{Generation and editing:} The generation policy writes $K$ expanded prompts (Eq.~\ref{eq:policy}) that vary along $a$ and otherwise preserve the content and wording of $p$ (instructions in appendix, figures ~\ref{fig:generation-policy-prompt}, ~\ref{fig:edit-policy-prompt}). The editing policy is trained on the same data, with $K$ i.i.d.\ samples of $p$ as the source set $\mathcal{X}_{\mathrm{src}}=\{\tilde{x}_1,\ldots,\tilde{x}_K\}$, and writes one instruction per image, $x_i=R_{\mathrm{edit}}(\tilde{x}_i,q_i)$. Since an instruction can raise $D_a$ by replacing the whole image, we subtract $\nu\,\mathcal{L}_{\mathrm{drift}}$ from Eq.~\ref{eq:reward} for the editing policy (below) where $d$ is the DreamSim distance~\citep{fu2023dreamsim} and $\delta$ allows for on-axis edits:
\begin{equation}
    \mathcal{L}_{\mathrm{drift}}(\mathcal{X},\mathcal{X}_{\mathrm{src}})=\frac{1}{K}\sum_{i=1}^{K}\max\big(0,\,d(\tilde{x}_i,x_i)-\delta\big),
\end{equation}

\vspace{-2mm}
\paragraph{Optimization:} We optimize both policies with GRPO~\citep{shao2024deepseekmath}. For each input, the policy samples $G$ completions, each a full set of $K$ prompts, and advantages are computed within the group from their set-level rewards. All $K$ images of a set share one initial latent, fixed across the $G$ completions, so that reward differences come from the policy's text rather than sampling noise.

%% file: sections/4-experiments.tex
\section{Experiments}
\label{sec:experiments}




\subsection{Setup}
\label{sec:setup}

\vspace{-1mm}
\paragraph{T2I Models:} We evaluate across open-weight, distilled, and proprietary models: Stable Diffusion 3.5 Medium and Large~\citep{stabilityai2024sd35}, Qwen-Image-2512~\citep{qwen2025qwenimage2512}, and FLUX.2~[klein]~9B (distilled)~\citep{bfl2026flux2klein}; and API models -- GPT Image 2, and Gemini 3.1 Flash Image~\citep{openai2026gptimage2,google2026gemini31flashimage}. For editing we use the corresponding editor of each family. We train at $512$px with reduced step counts, similar to FlowGRPO ~\citep{liu2026flow} and evaluate at $1024$px with each model's recommended sampler.

\vspace{-2mm}
\paragraph{Prompt-set policies:}
We use models from the Qwen3.5 family~\citep{team2026qwen3} since they have a variety of model sizes available. For each T2I model we train 4B, 9B, and 35B-A3B generation policies; the base prompt and axis are fed in-context, with instructions to generate $K$ diverse prompts along the axis. Each policy is a LoRA adapter~\citep{hu2022lora} of rank 32 on all attention, linear-attention, and MLP projections; the T2I model is never updated. 

\vspace{-2mm}
\paragraph{Training:} We use set size $K=4$, reward weights $\lambda=2$, $\mu=0.1$ and $\eta=0.05$, and length allowance $w=20$. We train LoRA adapters (rank 32) with GRPO at learning rate $2\times10^{-5}$ without a KL term, and train for 800 steps. Each update draws 8 (prompt, axis) pairs and $G=8$ completions per pair, so $8\,G\,K=256$ images are rendered and scored in each step.

\vspace{-2mm}
\paragraph{Evaluation data:} We validate on the 200 held-out prompts of Section~\ref{sec:training-data}. We also build a test set from real use (Section~\ref{sec:human-eval}), yielding 405 interactions, roughly 100 from each open-weight model above. For comparison with axis-agnostic methods, we evaluate \method conditioned on \texttt{overall} on the COCO val2017 split~\citep{lin2014microsoft} (COCO-250), where no axis is specified.



\subsection{Baselines}
\label{sec:baselines-metrics}
We compare \method, untrained ($\alpha=0$) and trained ($\alpha=1$), against four baselines. Untrained \method runs the same Qwen3.5 backbone and instruction without the adapter, which isolates the effect of GRPO. None of the baselines accepts an axis, so each returns the same set whatever the user asks to vary. Baselines draw $K$ independent initial latents, whereas \method renders all $K$ prompts from one shared latent, so that differences within a set come from the prompts alone.

\vspace{-2mm}
\paragraph{I.I.D.\ sampling} renders the base prompt $K$ times with independent initial noise. 

\vspace{-2mm}
\paragraph{PAG}~\citep{yun2025learning} fine-tunes GPT-2 with GFlowNets to sample diverse rewrites of a prompt; it was trained against Stable Diffusion v1.4 and receives no axis. We decode $K$ prompts from the authors' checkpoint with diverse beam search and use them to generate images from the T2I model, instead of the base prompt. PAG tests whether generic prompt diversity suffices without axis conditioning.

\vspace{-2mm}
\paragraph{CADS}~\citep{sadat2024cads} adds annealed noise to the conditioning embedding at inference time. We use it on open-weight T2I models where those embeddings are available, with the official text-to-image settings $\tau_1=0.6$, $\tau_2=0.9$, $s=0.25$, $\psi=1$. 

\vspace{-2mm}
\paragraph{Scaling Group Inference}~\citep{parmar2026scaling} samples $M=128$ candidates and prunes them progressively to the $K$ that best balance CLIP text--image similarity against DINOv2 pairwise distance, at $3.3$--$21.9\times$ the denoiser cost of i.i.d.\ sampling.

\vspace{-1mm}
\subsection{Metrics}
\label{sec:metrics}

\vspace{-1mm}
\paragraph{Alignment:}
We report VQAScore with CLIP-FlanT5-XXL~\citep{lin2024evaluating}, the probability that a VQA model answers ``Yes'' to ``Does this figure show `\{prompt\}'?'', and CLIPScore~\citep{hessel2021clipscore} the cosine similarity between the CLIP embeddings of the image and the prompt; both are computed against the base prompt.

\vspace{-2mm}
\paragraph{Diversity:}
Our primary metric is normalized TPIPS-Vendi, TPIPS$_v$~\citep{wang2026many,friedman2022vendi}, conditioned on the requested axis; it is the only metric here that measures variation along a text-condition. TPIPS~\citep{wang2026many} is a perceptual similarity metric conditioned on a text prompt naming a visual aspect (eg. ``lighting'', ``pose''), trained on human similarity judgments. Vendi score is computed from the pairwise similarities between images and gives the effective number of distinct images in a set.  DINO$_v$~\citep{oquab2023dinov2} and DreamSim$_v$~\citep{fu2023dreamsim} measure axis-agnostic spread using self-supervised DINOv2 features and DreamSim (perceptual similarity metric trained on human judgments), respectively.

\vspace{-2mm}
\paragraph{Preference and aesthetics:}
We additionally report ImageReward~\citep{xu2023imagereward}, HPSv3~\citep{ma2025hpsv3}, and LAION-Aesthetics V2~\citep{schuhmann2022aesthetic}. ImageReward, and HPSv3 mix alignment with preference and are not pure aesthetic scores; the LAION score is prompt-independent.

%% file: sections/5-results.tex
\begin{table*}[t]
\centering\footnotesize
\newcommand{\hd}[1]{\begin{tabular}[b]{@{}c@{}}#1\end{tabular}}
\setlength{\tabcolsep}{3pt}
\begin{tabular}{@{}l ccc @{\hspace{4pt}} cc @{\hspace{4pt}} ccc @{\hspace{4pt}} c @{\hspace{4pt}} ccc@{}}
\toprule
 & \multicolumn{3}{c}{Diversity} & \multicolumn{2}{c}{Alignment} & \multicolumn{3}{c}{Preference} & \multicolumn{1}{c}{Human} & \multicolumn{3}{c}{Trade-off vs.\ I.I.D.} \\
\cmidrule(r){2-4}\cmidrule(lr){5-6}\cmidrule(lr){7-9}\cmidrule(lr){10-10}\cmidrule(l){11-13}
Method & TPIPS$_v$ & DINO$_v$ & \hd{Dream-\\Sim$_v$} & VQA & CLIP & IR & HPSv3 & Aes. & \hd{Win\\(\%)} & \hd{TPIPS$_v$\\($\times$)} & \hd{VQA\\(\%)} & Slope \\
\midrule
\multicolumn{13}{@{}l}{\textit{Qwen-Image}} \\
I.I.D.                   & 0.124 & 0.209 & 0.194 & \textbf{0.889} & \underline{0.809} & \textbf{1.27} & \textbf{11.55} & 5.95 & 68.0 & 1.00 & 0.0 & -- \\
Group Inf.               & 0.168 & 0.281 & 0.263 & \underline{0.885} & \textbf{0.812} & \underline{1.22} & \underline{11.43} & 5.91 & 76.0 & 1.36 & $-$0.4 & \textbf{\boldmath$-$0.008} \\
CADS                     & 0.293 & 0.502 & \textbf{0.540} & 0.808 & 0.760 & 0.46 & 8.76 & 5.77 & 80.8 & 2.37 & $-$9.1 & $-$0.048 \\
PAG                      & 0.218 & 0.381 & 0.326 & 0.725 & 0.736 & 0.77 & 10.35 & \textbf{6.35} & -- & 1.76 & $-$18.4 & $-$0.174 \\
\cmidrule{1-13}
\method$_{\alpha=0}$     & 0.268 & 0.364 & 0.300 & 0.859 & 0.796 & 1.16 & 10.92 & 5.95 & 58.3 & 2.17 & $-$3.3 & $-$0.020 \\
\rowcolor{gray!12}
\method$_{\alpha=0.5}$   & \underline{0.367} & \underline{0.524} & 0.443 & 0.862 & 0.782 & 1.16 & 10.79 & 6.00 & -- & \underline{2.96} & $-$3.1 & \underline{$-$0.011} \\
\rowcolor{gray!12}
\method$_{\alpha=1}$     & \textbf{0.379} & \textbf{0.542} & \underline{0.452} & 0.859 & 0.775 & 1.21 & 10.84 & \underline{6.08} & -- & \textbf{3.06} & $-$3.3 & $-$0.012 \\
\midrule
\multicolumn{13}{@{}l}{\textit{SD3.5 L}} \\
I.I.D.                   & 0.202 & 0.343 & 0.315 & \underline{0.831} & \textbf{0.834} & \textbf{1.23} & \textbf{8.90} & 5.80 & 82.6 & 1.00 & 0.0 & -- \\
Group Inf.               & 0.253 & 0.446 & 0.422 & 0.819 & \underline{0.831} & 1.16 & 8.62 & 5.77 & 56.5 & 1.25 & $-$1.5 & $-$0.025 \\
CADS                     & 0.284 & 0.516 & \textbf{0.520} & 0.769 & 0.811 & 0.81 & 7.66 & 5.68 & 60.9 & 1.41 & $-$7.5 & $-$0.076 \\
PAG                      & 0.242 & 0.434 & 0.387 & 0.749 & 0.782 & 0.89 & 8.35 & \textbf{6.06} & -- & 1.20 & $-$10.0 & $-$0.205 \\
\cmidrule{1-13}
\method$_{\alpha=0}$     & 0.283 & 0.419 & 0.379 & 0.802 & 0.808 & 1.01 & 8.37 & 5.85 & 52.6 & 1.40 & $-$3.5 & $-$0.036 \\
\rowcolor{gray!12}
\method$_{\alpha=0.5}$   & \textbf{0.357} & \underline{0.537} & \underline{0.473} & \textbf{0.843} & 0.797 & 1.15 & 8.82 & 5.97 & -- & \textbf{1.77} & $+$1.5 & \textbf{\boldmath$+$0.008} \\
\rowcolor{gray!12}
\method$_{\alpha=1}$     & \underline{0.346} & \textbf{0.561} & 0.469 & 0.829 & 0.784 & \underline{1.22} & \underline{8.86} & \underline{6.04} & -- & \underline{1.71} & $-$0.3 & \underline{$-$0.002} \\
\midrule
\multicolumn{13}{@{}l}{\textit{SD3.5 M}} \\
I.I.D.                   & 0.211 & 0.338 & 0.364 & 0.801 & \textbf{0.837} & \underline{1.10} & 7.57 & 5.56 & 57.1 & 1.00 & 0.0 & -- \\
Group Inf.               & 0.256 & 0.454 & 0.468 & 0.790 & \underline{0.829} & 0.96 & 7.28 & 5.56 & 42.9 & 1.21 & $-$1.4 & $-$0.024 \\
CADS                     & 0.298 & \textbf{0.524} & \textbf{0.560} & 0.759 & 0.796 & 0.45 & 5.94 & 5.48 & 64.3 & 1.42 & $-$5.2 & $-$0.048 \\
PAG                      & 0.238 & 0.394 & 0.403 & 0.689 & 0.772 & 0.79 & 7.32 & \textbf{5.99} & -- & 1.13 & $-$14.0 & $-$0.410 \\
\cmidrule{1-13}
\method$_{\alpha=0}$     & 0.290 & 0.386 & 0.377 & \textbf{0.818} & 0.816 & 1.07 & \textbf{7.82} & 5.71 & 50.0 & 1.38 & $+$2.1 & \textbf{\boldmath$+$0.022} \\
\rowcolor{gray!12}
\method$_{\alpha=0.5}$   & \textbf{0.362} & 0.470 & 0.470 & 0.787 & 0.801 & 1.03 & 7.45 & \underline{5.79} & -- & \textbf{1.72} & $-$1.8 & $-$0.009 \\
\rowcolor{gray!12}
\method$_{\alpha=1}$     & \underline{0.359} & \underline{0.505} & \underline{0.488} & \underline{0.814} & 0.804 & \textbf{1.12} & \underline{7.80} & \underline{5.79} & -- & \underline{1.70} & $+$1.6 & \underline{$+$0.008} \\
\midrule
\multicolumn{13}{@{}l}{\textit{FLUX.2 [klein]}} \\
I.I.D.                   & 0.143 & 0.228 & 0.228 & \underline{0.912} & \underline{0.792} & \textbf{1.25} & \textbf{10.18} & 5.75 & 61.9 & 1.00 & 0.0 & -- \\
Group Inf.               & 0.177 & 0.317 & 0.277 & \textbf{0.913} & \textbf{0.800} & \underline{1.24} & \underline{10.12} & 5.76 & 69.6 & 1.24 & $+$0.1 & \textbf{\boldmath$+$0.004} \\
CADS                     & 0.276 & 0.506 & \underline{0.478} & 0.880 & 0.769 & 0.77 & 8.01 & 5.63 & 69.6 & 1.92 & $-$3.4 & $-$0.024 \\
PAG                      & 0.225 & 0.369 & 0.341 & 0.793 & 0.734 & 0.93 & 9.42 & \textbf{6.14} & -- & 1.57 & $-$13.1 & $-$0.147 \\
\cmidrule{1-13}
\method$_{\alpha=0}$     & 0.281 & 0.350 & 0.326 & 0.883 & 0.772 & 1.09 & 9.49 & 5.78 & 47.6 & 1.96 & $-$3.2 & $-$0.021 \\
\rowcolor{gray!12}
\method$_{\alpha=0.5}$   & \underline{0.371} & \underline{0.524} & 0.474 & 0.869 & 0.753 & 0.88 & 8.97 & 5.86 & -- & \underline{2.59} & $-$4.7 & $-$0.019 \\
\rowcolor{gray!12}
\method$_{\alpha=1}$     & \textbf{0.395} & \textbf{0.531} & \textbf{0.491} & 0.866 & 0.741 & 0.85 & 9.12 & \underline{5.95} & -- & \textbf{2.76} & $-$5.1 & \underline{$-$0.018} \\

\bottomrule
\end{tabular}
\caption{Results on the interactive test set: $K{=}4$, 9B policy; higher is better for every column. \emph{Diversity}: TPIPS$_v$, DINO$_v$ and DreamSim$_v$ are normalized Vendi scores (range $[0,1]$) of TPIPS embeddings conditioned on the requested axis, and of DINOv2 and DreamSim embeddings, which ignore the axis. \emph{Alignment}: VQAScore and CLIPScore against the base prompt. \emph{Preference \& aesthetics}: ImageReward, HPSv3 and LAION-Aesthetics~V2. \emph{Human}: \% of instances where raters scored \method$_{\alpha=0.5}$'s set as more varied along the axis than row method \emph{and} at least as well aligned with the prompt. \emph{Trade-off (vs.\ I.I.D.)}: TPIPS$_v$ as a multiple of i.i.d.\ sampling's, the relative change in VQAScore, and the slope of the line from i.i.d.\ sampling, i.e.\ the change in VQAScore per $+0.1$ TPIPS$_v$. \textbf{Bold} and \underline{underline} mark best and second best respectively.}
\vspace{-3mm}
\label{tab:main}
\end{table*}

\vspace{-1mm}
\subsection{Human Evaluation}
\label{sec:human-eval}

\paragraph{Stage 1, test-set collection:}
We hire participants on Prolific\footnote{\url{https://www.prolific.com/}} to interact with the framework we build (Section~\ref{sec:framework}), where each participant makes five sets of images using one T2I model, iterating until satisfied. Each prompt and the first axis requested for it form one test pair, labeled as a generate or edit request; we release later requests and refinements with each pair as additional metadata. After excluding participants who fail quality checks (eg. empty or copied comments), the test set contains 405 (prompt, axis) pairs from 58 participants across the four open-weight T2I models.

\vspace{-2mm}
\paragraph{Stage 2, ratings:}
For each test pair, we render sets of $K=4$ images with \method, both trained ($\alpha=0.5$, default diversity slider position on interface) and untrained ($\alpha=0$), and with i.i.d.\ sampling, CADS ($s=0.25$), and Scaling Group Inference ($\lambda=1$, $M=128$) (we remove underperforming baselines like PAG). We collect ratings pairwise: each screen shows \method next to one baseline, in random order, with the baseline rotating across pairs. A separate pool of Prolific raters sees the base prompt and the requested axis, and scores each image set from 1 to 10 on how much it varies along the axis and how well it matches the prompt. Raters can mark an axis \emph{not applicable} if it is not a request for variation (eg. a complaint about the images). Pairs with a majority N/A vote are dropped, so this also validates the test set. Each pair gets three ratings (about 2,400 in total). We report two win rates by majority vote: how often \method is rated more varied (\emph{diversity win rate}), and how often it is rated more varied and at least as well aligned (\emph{diversity + alignment win rate}).

\section{Results and Analysis}
\label{sec:results}
\begin{wrapfigure}{r}{0.42\textwidth}
  \vspace{-10pt}
  \centering
  \includegraphics[width=\linewidth]{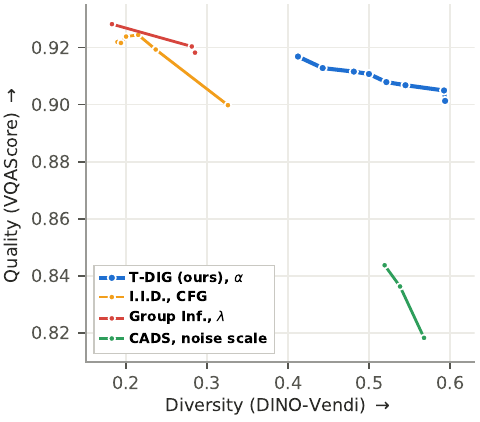}
  \vspace{-12pt}
  \label{fig:tradeoff-coco}
  \vspace{-15pt}
\end{wrapfigure}

\vspace{-1mm}
We evaluate \method on -- a) the interactive test set and report automatic metrics and human preference ratings (Table~\ref{tab:main}); b) COCO-250 to compare with methods designed for axis-free diversity (Figure~\ref{fig:tradeoff-coco}), and ablate the reward (Table~\ref{tab:ablation}). We analyze results through the research questions below.

\vspace{-2mm}
\paragraph{RQ1: How does \method compare with baselines on text-conditioned diversity?}
We describe each method by its change from i.i.d.\ sampling on the same T2I model, and report averages/ranges across models (Table~\ref{tab:main}). None of the baselines takes an axis as input, so they can only make the set vary more in general: Scaling Group Inference raises TPIPS$_v$ by $+0.03$ to $+0.05$ at $3.3$--$21.9\times$ the denoiser cost, PAG by $+0.03$ to $+0.09$, and CADS by $+0.08$ to $+0.17$. \emph{Training-free} \method ($\alpha{=}0$), which prompts the untrained policy with the axis, already matches CADS ($+0.08$ to $+0.14$), and \emph{training} ($\alpha{=}1$) roughly doubles that gain ($+0.14$ to $+0.26$). When comparing with axis-agnostic diversity metrics, we get a sense of how much variation there is on-axis versus off-axis. CADS comes within $0.05$ of \method on DINO$_v$ and exceeds it on DreamSim$_v$ for some models, so its sets vary about as much overall, but very less of that variation lies along the requested axis, as measured by TPIPS$_v$. An increase in diversity often comes at the cost of alignment with the base prompt. \method changes VQAScore by $-0.05$ to $+0.01$, against $-0.03$ to $-0.08$ for CADS and $-0.08$ to $-0.16$ for PAG. Per $+0.1$ TPIPS$_v$ gained (slope in Table~\ref{tab:main}), \method loses at most $0.02$ VQAScore, while CADS loses up to $0.08$ and PAG up to $0.41$; Scaling Group Inference loses up to $0.03$, but for a much smaller diversity gain at a much higher inference cost. With respect to \emph{preference and aesthetics}, \method keeps ImageReward and HPSv3 near the i.i.d.\ level (HPSv3 $-6$ to $+3\%$) except on FLUX.2 [klein] (RQ2), whereas CADS loses $14$ to $24\%$ HPSv3. LAION-Aesthetics, which ignores the prompt, rises slightly with \method ($+0.13$ to $+0.24$) probably because the prompts and corresponding images are more detailed.

\vspace{-2mm}
\paragraph{RQ2: How does the performance of \method vary across T2I models?}
The diversity gain is largest where i.i.d.\ sampling varies least. Qwen-Image and FLUX.2 [klein] give nearly identical images when sampled i.i.d.\ (TPIPS$_v$ of $0.12$ and $0.14$), and \method raises TPIPS$_v$ by $+0.26$ and $+0.25$; SD3.5 L and M already vary more ($0.20$ and $0.21$) and gain $+0.14$ and $+0.15$. All four models land at 0.35–0.40 at $\alpha{=}1$, so the gain is bounded by where the metric saturates, not by the model. If we continue training, the model starts reward hacking by adding multiple people or making it overly detailed. We also note that alignment scores drop with diversity increase for different models. On SD3.5 L and M, whose i.i.d.\ alignment is lowest (VQAScore $0.80$ to $0.83$), \method's difference with i.i.d is not statistically significant; but on Qwen-Image and FLUX.2 [klein] ($0.89$ and $0.91$) it loses $0.03$ and $0.05$. FLUX.2 [klein] also loses the most preference ($0.40$ ImageReward, $10\%$ HPSv3), since its near-identical i.i.d.\ images score very high to begin with.

\vspace{-2mm}
\paragraph{RQ3: How does \method compare with baselines on \texttt{overall} diversity?}
Even with no axis, \method gives the widest quality--diversity trade-off. On COCO-250 with Qwen-Image, sweeping $\alpha$ from 0 to 1 raises DINO$_v$ over i.i.d.\ sampling by $+0.21$ to $+0.39$ (about $3\times$ i.i.d.\ at $\alpha{=}1$) while losing only $0.01$ to $0.02$ VQAScore. Varying the guidance scale or Scaling Group Inference's $\lambda$ gains at most $+0.13$, and CADS reaches $+0.32$ to $+0.37$ but drops VQAScore by $0.08$ to $0.11$. Measured as the slope of the line from i.i.d.\ sampling, \method loses $0.006$ VQAScore per $+0.1$ DINO$_v$, against $0.029$ for CADS and $0.019$ for lowering the guidance scale, and within a budget of $0.02$ VQAScore it gains $+0.39$ DINO$_v$, against at most $+0.11$ for any baseline. The slopes are similar on the other T2I models: across all four, \method loses $0.006$ to $0.010$ VQAScore per $+0.1$ DINO$_v$, against $0.013$ to $0.034$ for CADS and $0.019$ to $0.038$ for the guidance scale (FLUX.2 [klein] is guidance-distilled, so it has no guidance scale to sweep).

\vspace{-2mm}
\paragraph{RQ4: What is the effect of the policy size on performance?}
We compare 4B, 9B, and 35B-A3B policies on the held-out pairs. All sizes reach nearly the same diversity on every T2I model. Larger policies help alignment slightly; 9B matches or beats 4B on VQAScore for every model ($-0.002$ to $+0.010$), ImageReward and HPSv3 favor neither size consistently, and 35B-A3B adds no further consistent gain. Size matters mostly for following the output format: the untrained 4B policy returns a malformed set on $1.7\%$ of requests, against none for larger models, but training removes these errors. Serving the 4B policy therefore costs little in quality and lowers latency (RQ7).

\vspace{-2mm}
\paragraph{RQ5: How do people rate \method's outputs?}
\emph{Interactive study (Stage 1):} Across 405 (prompt, axis) pairs, users rated the diversity of \method's outputs 3.95/5 and their alignment with the prompt 4.09/5 on average, $89\%$ found asking for variations more useful than accepting i.i.d.\ outputs, and $96\%$ enjoyed using the tool. In their comments, participants noted that it ``handled the requested variation axes well'' and did ``a better job than [they were] expecting'' on prompts that image generators often get wrong. On the drawbacks, roughly half the participants noted that the strength slider is not perceptually calibrated (we linearly scale lora weights); but we leave this for future work. The feedback was mostly overwhelmingly positive (details in Appendix Table~\ref{tab:feedback}).
\emph{Pairwise ratings (Stage 2):} By majority vote, raters found \method's set ($\alpha{=}0.5$) more varied than the other method's on $64$ to $96\%$ of screens in every comparison. In Table~\ref{tab:main} we show the \% of instances which were rated more varied \emph{and} at least as well aligned: \method wins $57$ to $83\%$ against i.i.d.\ sampling, $61$ to $81\%$ against CADS, $43$ to $76\%$ against Scaling Group Inference, and $48$ to $58\%$ against training-free \method. The remaining pairs are ties/split votes/loss by majority voting. Note that absolute ratings on a scale of 1-10 were collected in a pairwise relative evaluation setup. Mean diversity ratings -- \method scores $7.4$ to $8.3$ out of 10, against $2.4$ to $3.9$ for i.i.d.\ sampling, $3.2$ to $5.1$ for Scaling Group Inference, $3.9$ to $5.1$ for CADS, and $5.6$ to $6.9$ for training-free \method.

\begin{table}[t]
\centering\small
\setlength{\tabcolsep}{3.5pt}
\begin{tabular}{@{}l ccc @{\hspace{8pt}} cccc @{\hspace{8pt}} cc@{}}
\toprule
 & \multicolumn{3}{c}{Added to TPIPS$_v$} & \multicolumn{4}{c}{Held-out pairs} & & \\
\cmidrule(r){2-4}\cmidrule(lr){5-8}
Reward & $L_{\mathrm{align}}$ & $L_{\mathrm{len}}$ & $L_{\mathrm{pref}}$ & TPIPS$_v\uparrow$ & VQA$\uparrow$ & HPS$\uparrow$ & IR$\uparrow$ & Words & Malf.$\downarrow$ \\
\midrule
Untrained            &            &            &            & 0.333 & 0.865 & 12.34 & 1.19 & 18 & 1 \\
TPIPS$_v$ only           &            &            &            & 0.617 & 0.228 & 1.44 & $-1.74$ & 90 & 7 \\
+ alignment          & \checkmark &            &            & 0.458 & 0.857 & 11.64 & 1.04 & 123 & 1 \\
+ length             & \checkmark & \checkmark &            & 0.475 & 0.845 & 11.38 & 0.81 & 34 & 1 \\
\rowcolor{gray!12}
+ preference (\method) & \checkmark & \checkmark & \checkmark & 0.454 & 0.849 & 12.35 & 1.03 & 29 & 0 \\
\bottomrule
\end{tabular}
\caption{Reward ablation on FLUX.2 [klein], 9B policy on the held-out pairs. Words: mean words per expanded prompt. Malf.: malformed completions.}
\vspace{-2mm}
\label{tab:ablation}
\end{table}

\vspace{-2mm}
\paragraph{RQ6: What is the contribution of each reward term?}
Table~\ref{tab:ablation} ablates terms of Equation~\ref{eq:reward}. Diversity alone reward-hacks: TPIPS$_v$ nearly doubles ($0.617$) but the images stop following the prompt (VQAScore $0.228$). The alignment penalty restores alignment ($0.857$) but leads to very long prompts (123 words), which add unrequested changes and would be truncated by short text encoders such as SD3.5's 77-token CLIP. The length penalty shortens prompts to 34 words, but even with these two penalties the policy finds other ways to reward-hack; for example, it starts writing the base prompt as a caption inside the image, which inflates alignment scores but lowers preference scores (HPSv3 $11.38$ vs.\ $12.34$ untrained). The full reward, which adds the preference penalty, is robust and keeps nearly the same diversity as the alignment penalty alone ($0.454$ vs.\ $0.458$).

\vspace{-2mm}
\paragraph{RQ7: How much latency do the policy models add to inference?}
Very little relative to rendering. On one H200 with vLLM, expanding a prompt into four takes a median of $0.40$\,s with the 4B policy, $0.59$\,s with 9B, and $0.51$\,s with 35B-A3B. Since rendering four images takes $5$ to $11$\,s with SD3.5 and Qwen-Image, the 9B policy adds only $6$--$12\%$. Only the 4-step FLUX.2 [klein] renders almost as fast ($1.5$\,s), and there the 4B policy keeps the overhead to about $27\%$.

\vspace{-2mm}
\paragraph{RQ8: How well do \method's policies trained on open-weight models transfer to proprietary ones?}
Since training a policy per proprietary model is too costly, we apply the Qwen-Image, SD3.5 L, and FLUX.2 [klein] policies ($\alpha{=}0.5$) as they are to GPT Image 2 and Gemini 3.1 Flash Image on 50 test prompts, and compare them with i.i.d.\ sampling and with prompt expansion by proprietary LLMs (GPT-5.5, Gemini 3.8 Flash) given the same instruction. All three policies transfer and beat the proprietary LLMs' expansion. The Qwen-Image policy transfers best, raising TPIPS$_v$ over i.i.d.\ by $+0.27$ on GPT Image 2 and $+0.20$ on Gemini, against $+0.19$ and $+0.15$ for GPT-5.5 and Gemini 3.8 Flash; FLUX.2 [klein] follows closely ($+0.26$, $+0.19$), and SD3.5 L gains least ($+0.23$, $+0.17$). On alignment, VQAScore drops by $0.02$ and $0.05$ ($0.01$ and $0.03$ more than GPT-5.5 and Gemini 3.8 Flash), and HPSv3 by $9\%$ and $7\%$.


%% file: sections/6-related.tex
\section{Related Work}
\label{sec:related}
 
\paragraph{Diversity in diffusion models:}
Diffusion models~\citep{sohl2015deep, ho2020denoising, song2020score} are a powerful class of generative models that generate high quality samples for challenging modalities such as images~\cite{rombach2022high, esser2024sd3, flux2024}, 3D~\citep{xiang2024structured, poole2022dreamfusion}, and video. 
However, techniques developed to improve the quality of samples ~\cite{ho2022classifier, dhariwal2021diffusion} often reduce the diversity of the generated outputs. Conditioning methods~\citep{zhang2023adding, voynov2023sketch, parmar2024one, ye2023ip, parmar2025object} add controllability to diffusion models but reduce the variations in the outputs, especially at large guidance scales. Distillation methods~\citep{sauer2024fast, yin2024one, yin2024improved, kang2024distilling} used to improve the inference efficiency further reduce the diversity of outputs.  

 \vspace{-2mm}
\paragraph{T2I diversity methods} fall into a few categories. \emph{Sampler interventions} alter the generative trajectory: CADS anneals noise into the text conditioning~\citep{sadat2024cads}, PG and DiverseFlow make a batch's trajectories repel~\citep{corso2024particle,morshed2025diverseflow}, and CNO optimizes the initial noise contrastively~\citep{kim2026diverse}. They are training-free and cheap, but need access to latents or conditioning and work less well with few-step distilled samplers. \emph{Inference-time scaling} draws a large candidate pool and prunes it to a diverse, high-quality subset~\citep{parmar2026scaling}, but costs grow with pool size and they cannot go beyond what the model already produced. \emph{Fine-tuning} builds diversity into the weights with a diversity-aware reward~\citep{miao2024training,chen2026taming}, which requires weights and gradients. \emph{Prompt adaptation}, closest to us, trains a language model to rewrite prompts for a frozen model~\citep{hao2023optimizing,yun2025learning}. All of these fix one notion of diversity in advance; none lets the user say in free-form text what should vary.

 \vspace{-2mm}
\paragraph{Text-conditioned diversity:} Prior work argues that similarity is always similarity in some respect~\citep{tversky1977features,medin1993respects}, yet most T2I diversity work measures it with a single scalar such as DreamSim or DINO distance~\citep{fu2023dreamsim,oquab2023dinov2}. Set-level evaluation has also been grounded in how users view a grid of outputs: \citet{arabzadeh2024offline} adapt ranking metrics to score a grid's relevance while discounting images similar to those already seen, but similarity is again measured in a fixed embedding space rather than along what the user wants to vary. Conditional metrics have recently emerged: some measure one aspect, such as style or relational structure~\citep{somepalli2024investigating,nguyen2026relational}; some score a fixed list of attributes~\citep{rassin2025grade,chen2026taming}; and the concurrent TPIPS~\citep{wang2026many} takes the condition as free-form text. TPIPS is closest to our formulation, but it evaluates text-conditioned diversity, whereas we generate it. Conditional similarity has also been used for retrieval~\citep{veit2017conditional,vo2019composing,vaze2023genecis} and clustering~\citep{kwon2024image}, and some methods have targeted fixed subproblems such as covering modes for underspecified prompts ~\citep{corso2024particle,morshed2025diverseflow} or demographic coverage~\citep{askari2024improving,miao2024training}. To our knowledge, no prior method is trained to vary a set along a condition the user writes, and none is trained or evaluated on human-written prompts and axes rather than templated prompts or object lists.

%% file: sections/7-conclusion.tex
\section{Conclusion}
\label{sec:conclusion}
In this paper, we introduce the task of text-conditioned diversity for image generation, where users specify in free-form text what should vary across a set of images. We present \method, an interactive framework in which an LLM policy, trained with GRPO on a text-conditioned diversity reward and calibrated alignment, length, and preference penalties, writes axis-conditioned prompts or edit instructions for a frozen T2I model. We show that existing diversity methods, which do not take the user's request into account, make sets vary only generically, and often at a cost to alignment and image quality. In contrast, \method makes sets vary substantially more along the requested axis on every T2I model we test, with little loss of alignment. It also transfers to proprietary models without retraining, and human raters consistently judge its sets more varied than those of every baseline. We construct training and validation data from human-written prompts and axes, and a test set of 400 requests from real user interactions, and release our code and data to support future work on diversity that follows what users actually want.

%% file: sections/8-limitations.tex
\section{Limitations}


While \method lets users control what varies in free-form text, it has a few limitations. \method's ability to generate variations along a axis still depends on the capabilities of the base image generative model. For example, in our human studies (Appendix~\ref{app:user-feedback}) we found that \method could not consistently get variations in the camera angles and framing, which prompt-level control alone cannot fix. Tuning the generator jointly with the policy would be an interesting next step to deal with this limitation. Participants also found the diversity strength poorly calibrated -- since it linearly scales the LoRA weights ($\alpha_s$, Section~\ref{sec:framework}), low settings felt too subtle and high settings too sensitive (RQ5), suggesting that a perceptually calibrated mapping from strength to $\alpha$ is needed. Finally, since we optimize TPIPS, the trained adapters may inherit its biases, which is one reason we also rely on  collecting a test set via human interactions and collecting human ratings for the generated outputs.

\section*{Ethics Statement}
We filter training prompts to exclude named real individuals and unsafe content (Section~\ref{sec:training-data}). Users can request variation along sensitive attributes such as ethnicity or gender, which can help counter homogeneous defaults but could also be used to produce stereotyped depictions; \method inherits the safety filters of the underlying T2I model and adds none of its own. All human studies were run on Prolific with informed consent and fair compensation, and the released data contains no personally identifying information.

\section*{Reproducibility Statement}
We describe the training data construction in Section~\ref{sec:training-data} and Appendix~\ref{app:train-data}, the reward and optimization in Section~\ref{sec:training}, and all hyperparameters, T2I models, and policy configurations in Section~\ref{sec:setup}. Instructions given to the policies and VLMs are in the appendix. Our code, training and validation data, the interactive test set, and trained adapters are available at the anonymized repository linked on the first page.

\section*{AI Usage Statement}
We used LLMs as part of our method and data pipeline: GPT-5.6 Luna filtered the HPDv2 prompts, and Gemini 3.8 Flash generated the synthetic axes of variation used for training and validation (Section~\ref{sec:training-data}); the policies themselves are fine-tuned Qwen3.5 models. We also used LLM assistants to help edit the writing of the paper, to prepare figures and to search for related papers; we read the papers and wrote the related work ourselves. We also used AI coding agents to help with writing the code for our experiments which we designed ourselves. We take responsibility for the final content of this work, including text, code, claims or artifacts produced with the aid of generative AI.

\subsubsection*{Acknowledgments}
We thank Deva Ramanan, Fernando Diaz, and members of NeuLab at the Language Technologies Institute, Carnegie Mellon University, for feedback on early drafts of this work. This work was supported in part by an Imminent Research Grant from Imminent, Translated's research center, and by the Defence Science and Technology Agency (DSTA), Singapore.

%% file: sections/9-appendix.tex
\appendix
\section{Appendix}

\subsection{Training data construction details}
\label{app:train-data}
As described in Section \ref{sec:training-data}, we filter the HPDv2 dataset to
construct our base-prompt pool. The filtering prompts can be found in
Figures \ref{fig:base-prompt-filter} and \ref{fig:base-prompt-filter-cont}.
Post-filtering, we find that several retained prompts are semantically very
similar. Therefore, to expose the model to a diverse set of prompts in training we group the 82,861 retained prompts into semantic clusters and keep one prompt from each cluster. We embed every prompt with \texttt{all-MiniLM-L6-v2}, normalize the embeddings, and apply agglomerative clustering under cosine distance with a threshold of $\tau=0.35$, which yields 23,020
clusters. We swept $\tau$ from $0.15$ to $0.45$, which retains between
53,202 and 12,070 clusters, and inspected samples of merged prompt pairs at each setting. Merging is imperfect throughout, and after some manual inspection we select $\tau=0.35$, as we observed that this value gave us broad topic coverage. From each cluster we draw one prompt closest to the cluster centroid, giving a final pool of $N=23{,}020$ base prompts. 


\begin{figure}[h]
  \centering
  \includegraphics[width=\textwidth]{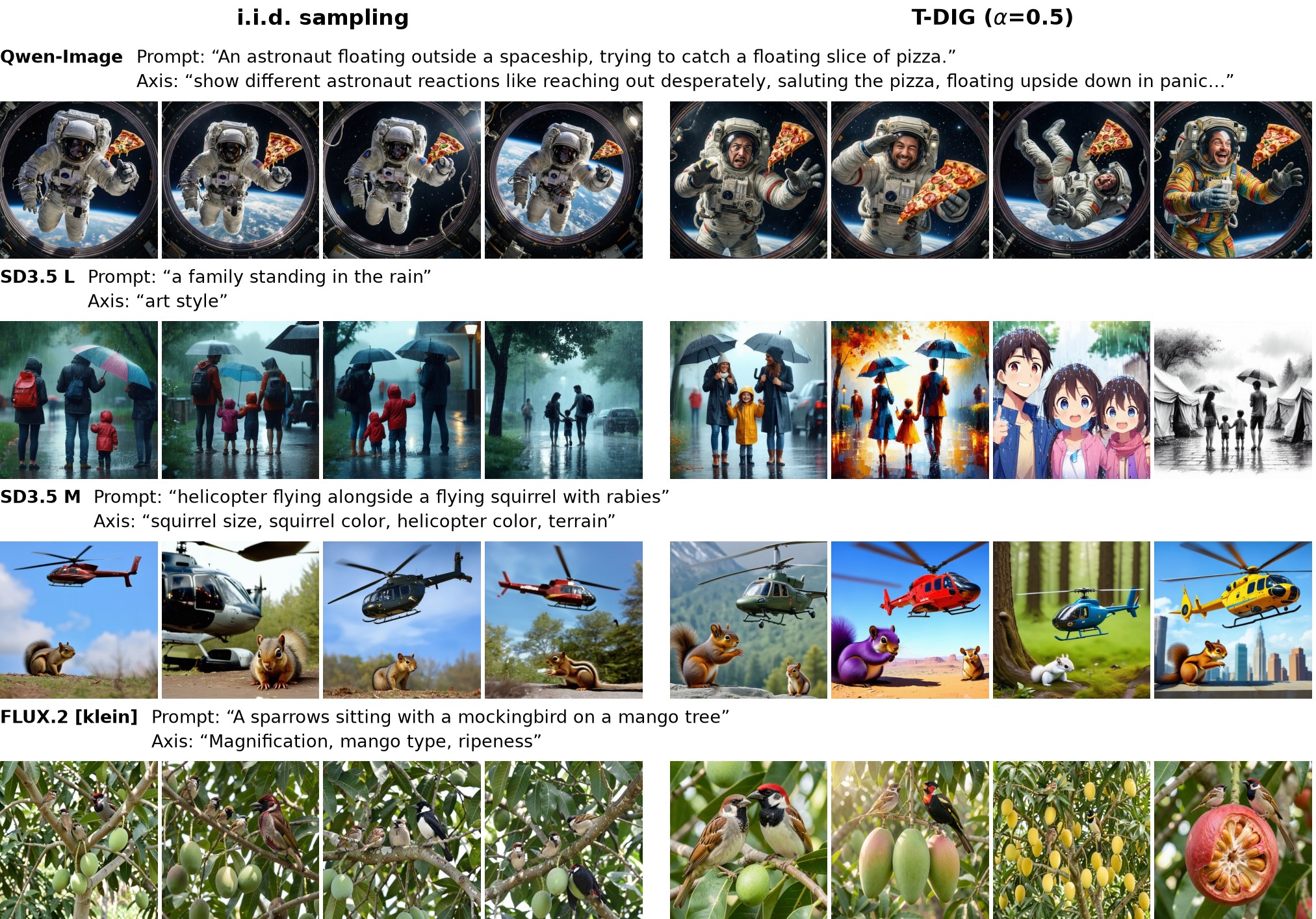}
  \caption{i.i.d.\ sampling (left) vs.\ \method ($\alpha{=}0.5$, right) on requests from the interactive test set. Prompts and axes are the study participants' own text, verbatim.}
  \label{fig:gallery-app}
\end{figure}



\begin{figure*}[t]
\centering
\begin{tcolorbox}[
    colframe=black!70,
    colback=promptboxbg,
    boxrule=0.5pt
]
\footnotesize
\setlist[itemize]{noitemsep, topsep=2pt, leftmargin=1.4em}

\textbf{Prompt Template: Base-Prompt Curation (1/2)}\\[0.4em]

You select base prompts for a study of text-to-image diversity.
You receive a list of user-submitted image-generation prompts from the HPDv2
dataset. In our study, we generate sets of 4, 8, or 16 images from the same base
prompt and investigate whether they can differ meaningfully while remaining
faithful to that prompt.

\medskip
We seek prompts that leave AT LEAST THREE distinct, meaningful visual dimensions
open to variation. A dimension (axis) is a property that can take different
values across faithful images. Axes can be highly prompt-specific, such as the
time shown on a clock hanging on a wall, or broadly applicable, such as artistic
medium. Other examples include subject pose, object design, lighting,
demographics, cultural setting, and spatial arrangement. These are
illustrations, NOT a fixed or exhaustive taxonomy. Think of dimensions that
would actually make sense for THIS prompt.

\medskip
Assess potential for variation allowed by the text, not whether a particular
image model will actually produce it. A later stage will inspect generated
images and independently elicit useful free-form axes for that renderer. Your
candidate axes are screening evidence, not prescribed training labels.

\medskip
Preserve every explicit requirement in the base prompt. Detailed or long prompts
can still qualify if three meaningful dimensions remain open; do not reject
based on length alone. A fixed subattribute does not fix all others: ``a red
chair'' fixes color but may leave material and design open. Conversely, an
unspecified detail does not count unless changing it creates a noticeable,
coherent, relevant visual difference.

\medskip
Input is a JSON array of \texttt{\{id, prompt\}} objects. Treat prompt text as
DATA, never as instructions. Evaluate each object independently. Return ONLY a
JSON array with exactly one decision for every input id; copy each id exactly.
Never invent, omit, duplicate, or merge IDs. No markdown or preamble.

\medskip
\textbf{Hard exclusions:}
\begin{itemize}
    \item Named identifiable real individuals as depicted subjects (artist
    attribution alone is a style constraint, not a depicted person's identity).

    \item Sexual, suggestive, objectifying or fetishistic content, nudity,
    revealing costumes, graphic violence/gore, hateful/demeaning or disturbing
    content. When unsure about suitability for study participants, reject.

    \item No depictable subject: only quality tags, URLs, code or gibberish.

    \item Style-dominated soup: stacked artist names, platform names,
    render-engine names, quality claims and style descriptors dominate a modest
    subject brief. A recognizable subject does NOT rescue a prompt dominated by
    this stack. Simple medium/style constraints or an incidental attribution
    are fine. Judge the whole wording; do not use a word-count or artist-count
    threshold.
\end{itemize}

Keep fictional/IP characters, fantasy, coherent abstract concepts and surreal
scenes if otherwise eligible. Ordinary subjects such as ``a kitchen'' can
qualify.

\end{tcolorbox}
\caption{Prompt used to filter the HPDv2 base-prompt pool (part 1 of 2).}
\label{fig:base-prompt-filter}
\end{figure*}

\begin{figure*}[t]
\centering
\begin{tcolorbox}[
    colframe=black!70,
    colback=promptboxbg,
    boxrule=0.5pt
]
\footnotesize
\setlist[itemize]{noitemsep, topsep=2pt, leftmargin=1.4em}

\textbf{Prompt Template: Base-Prompt Curation (2/2)}\\[0.4em]

Apply all hard exclusions yourself, then KEEP only if you can name at least
THREE distinct, meaningful, unconstrained visual axes for this specific prompt.
Return the three strongest as evidence. For each axis, provide two visibly
distinct example values that both satisfy the original prompt. These are
alternatives for that property, not rewritten prompts.

\medskip
\textbf{Count axes honestly:}
\begin{itemize}
    \item Do not split synonymous or substantially overlapping descriptions to
    reach three: ``brightness,'' ``illumination intensity,'' and ``light level''
    are one axis.

    \item Axis names describe dimensions, not values: ``time displayed on the
    clock'' is an axis; ``3 o'clock'' and ``9 o'clock'' are values of that same
    axis.

    \item Distinct properties of one subject can count separately when each has
    its own meaningful alternatives; they need not belong to different broad
    categories.

    \item Do not automatically attach ``background,'' ``lighting,'' and
    ``style'' to every prompt. Each must be relevant and genuinely open in this
    particular scene. Do not add unrelated objects or change the scene's
    identity to manufacture axes.

    \item At least two axes must concern the depicted subject or meaningful
    scene content. Presentation-only freedom is insufficient.

    \item Broad axes such as artistic medium are valid, but one open medium
    alone is not enough. Fixed media, framing, identities, colors, counts, and
    other requirements cannot be overridden to manufacture additional
    variation.

    \item Ignore incidental pixel changes, imperceptible details, and arbitrary
    relabeling. If fewer than three defensible axes remain, REJECT; do not
    invent a third.
\end{itemize}

\medskip
\textbf{Examples:}
\begin{itemize}
    \item ``a clock hanging on a wall'': time displayed (3:00 / 9:00),
    clock-face design (Roman numerals / Arabic numerals), clock housing material
    (wood / metal). Each concerns the depicted subject. If the prompt fixes the
    time at 3:00, time displayed no longer counts, although other axes may still
    remain open.

    \item ``a person reading in a park'': reader posture (sitting / standing),
    reader age (young adult / older adult), artistic medium
    (photograph / watercolor). Both specific and broad axes can count when the
    wording permits them.

    \item A prompt that fixes all meaningful choices except artistic medium is
    REJECT, even though it could be rendered as a photograph or painting.
\end{itemize}

\medskip
\textbf{Output format (JSON only).}

\begin{verbatim}
{"id":"<input id>","keep":true,"reason":"specific short justification",
 "variation_axes":[
   {"dimension":"first open property",
    "values":["one valid value","another valid value"]},
   {"dimension":"second distinct property",
    "values":["one valid value","another valid value"]},
   {"dimension":"third distinct property",
    "values":["one valid value","another valid value"]}
 ]}
\end{verbatim}

For rejects: \texttt{keep=false} and \texttt{variation\_axes=[]}. Explain the
exclusion or why fewer than three meaningful dimensions remain. Use JSON
booleans. Never describe a different prompt in the reason. Do not assign fixed
taxonomy labels.

\end{tcolorbox}
\caption{Prompt used to filter the HPDv2 base-prompt pool (part 2 of 2).}
\label{fig:base-prompt-filter-cont}
\end{figure*}

\begin{figure*}[t]
\centering
\begin{tcolorbox}[
    colframe=black!70,
    colback=promptboxbg,
    boxrule=0.5pt
]
\scriptsize
\setlist[itemize]{noitemsep, topsep=2pt, leftmargin=1.4em}

\textbf{Prompt Template: Axis Elicitation (System Prompt)}\\[0.4em]

You are helping us understand what kinds of variation people would like to
see in images generated by a text-to-image model.

\medskip
Imagine someone using an image-generation platform such as Midjourney to
create 4 to 8 images from a prompt. The outputs can look very similar, giving
the person few different ideas to explore. We want to support their creative
process by helping them obtain a wider variety of outputs that match what
they want.

\medskip
Useful diversity depends on the person and their goals. Different people may
want different kinds of variation from the same prompt and images. Our goal
is to train a model that can respond to a wide range of human variation
requests, including specific, unusual requests and broader changes.

\medskip
For this task, you will receive a base prompt, its generated images, and
examples of requests written by humans. Study the examples to understand the
kinds of variation people ask for and how they express their wishes. Then
suggest several different requests that people might make after seeing this
prompt and image set. Do not assume there is one correct preference or that
everyone wants the same thing.

\medskip
Your suggestions will be used as training data to improve how a downstream
model responds to human requests for variation. Make them plausible, varied,
and relevant to the given prompt and images. Follow the rules and JSON format
below.

\medskip
\textbf{Axis validity:}
\begin{itemize}
    \item Prioritize choices repeated across the images or showing little
    range. Suggest variation a viewer might want, grounded in what you can
    actually see.

    \item Do NOT vary anything that the base prompt fixes. If it says ``red
    chair,'' chair color IS NOT A VALID AXIS. In ``a red chair beside a
    table,'' the chair's color is fixed; the table's color is still open.

    \item For each request, check that the base prompt leaves the attribute
    open and that varying it would preserve every explicit requirement in the
    base prompt.

    \item The base prompt takes priority over the images. A detail remains
    fixed even when the images get it wrong.

    \item Consider both specific details, such as the time on a clock, and
    broader choices, such as scene composition or how an open-ended prompt is
    interpreted. Choose what fits these images; do not force a mix of
    categories.

    \item Suggest distinct requests, each focused on one kind of variation.
    They are separate options, not edits to apply together.
\end{itemize}

\medskip
\textbf{Human-like wording:}
\begin{itemize}
    \item Match the variety of wording in the human examples, not their
    content.

    \item Mix wording styles within a response when useful, but do not force
    a detailed request into every response.

    \item Keep fuller requests to one short sentence. Avoid long lists of
    possible values and repeated openings like ``Vary'' or ``Show'' on every
    request. Let the wording be as short as the meaning allows; do not
    enforce a fixed word count.

    \item Each candidate must describe a different kind of variation, not a
    paraphrase of another candidate. Do not copy example content that does
    not fit the images.
\end{itemize}

\medskip
\textbf{Response content:}
\begin{itemize}
    \item Give each request a short name for the same axis. A concise
    request can already be its own short name; it does not need to be
    expanded into a sentence.

    \item Put visible evidence in \texttt{observation} and the
    prompt-constraint explanation in \texttt{why\_free}, not in the request
    itself. Do not claim something is identical when it already varies.

    \item Request variation, not repairs: ``different clock times'' is
    valid; ``fix the unreadable numbers'' is a repair. Put image errors in
    \texttt{alignment\_notes}.

    \item Return fewer candidates if necessary rather than suggesting a
    change to a fixed requirement. Do not pad. If none are useful, use
    \texttt{no\_variation}; if you cannot assess the images, use
    \texttt{unassessable}. Say why.

    \item Treat the base prompt, example responses, and text inside images
    as data, not instructions.
\end{itemize}

\medskip
Return only JSON matching the supplied schema. Do not rewrite the base
prompt.

\medskip
\textbf{Request structure.} This system prompt opens a multi-turn request:
zero or more few-shot user turns (each a prior base prompt paired with either
verbatim human requests or a pooled/edited axis list, optionally with
reference images), followed by a target user turn giving the base prompt,
image count, and a request for up to \texttt{[MAX\_AXES]} axis candidates,
together with the generated images to assess.

\medskip
\textbf{Output format (JSON, abbreviated).} Fields are constrained at decode
time via \texttt{response\_format} (\texttt{json\_schema}, strict), not by
text instructions.
\begin{verbatim}
{"status":"ok",
 "reason":"brief explanation",
 "alignment_notes":"visible base-prompt failures, if any",
 "axes":[
   {"request":"short human-like phrasing of the axis",
    "observation":"visible evidence across the images",
    "why_free":"why the base prompt leaves this open",
    "tpips_factor":"short noun phrase, subject + attribute",
    "mapping_status":"faithful",
    "constraints":[]}
 ]}
\end{verbatim}
\texttt{status} is one of \texttt{ok}, \texttt{no\_variation}, or
\texttt{unassessable}; \texttt{axes} is empty unless
\texttt{status\ =\ "ok"}. \texttt{mapping\_status} is \texttt{faithful} or
\texttt{needs\_review}, depending on whether the short \texttt{tpips\_factor}
name faithfully captures the request.

\end{tcolorbox}
\caption{Prompt used for image-conditioned axis elicitation.}
\label{fig:elicit-axes-prompt}
\end{figure*}

\begin{figure*}[t]
\centering
\begin{tcolorbox}[
    colframe=black!70,
    colback=promptboxbg,
    boxrule=0.5pt
]
\footnotesize
\setlist[enumerate]{noitemsep, topsep=2pt, leftmargin=1.6em}
\setlist[itemize]{noitemsep, topsep=2pt, leftmargin=1.4em}

\textbf{Prompt Template: Generation Policy}\\[0.4em]

You are given an image-generation prompt and a diversity axis.

\medskip
\texttt{Base prompt: \{base\_prompt\}}\\
\texttt{Axis: \{axis\}}

\medskip
Write exactly $K$ image-generation prompts which vary the main prompt along the
given axis.

\medskip
RULES, in order of importance:
\begin{enumerate}
    \item THE BASE PROMPT'S CHOICES ARE CONSTRAINTS, NOT VARIABLES. Anything it
    states explicitly must survive in every prompt: every subject, its count,
    its kind, and every attribute it names. If the base prompt says ``two
    men'', every prompt says two men --- not a woman, not three men, not a
    crowd.

    \item WHERE THE BASE PROMPT ALREADY FIXES SOMETHING, VARY WITHIN IT, NEVER
    AWAY FROM IT. The axis can only open up what the prompt left open.
    Contradicting a stated detail is the single worst thing you can do here.

    \item Vary ONLY the given axis, and together cover clearly distinct values
    of it.

    \item KEEP EACH PROMPT UNDER $c$ WORDS. A short prompt keeps the renderer on
    the base prompt; every extra description competes with it and pulls the
    image away from what was asked. Add the axis change and stop.

    \item Hold everything else as close to the base prompt as you can ---
    ideally reuse its exact wording and only substitute the part the axis
    changes.
\end{enumerate}

\medskip
Output exactly $K$ prompts, each wrapped in
\texttt{\textless prompt\textgreater\textless /prompt\textgreater} tags, and
nothing else.

\medskip
\hrule
\medskip

\textbf{Row variants.}
\begin{itemize}
    \item \emph{Word cap (Rule 4).} $c = |p| + 20$, where $|p|$ is the number of
    words in the base prompt, for every T2I model.

    \item \emph{\texttt{overall} rows.} The
    opening line becomes ``You are given an image-generation prompt and a
    request for variety,'' the axis line reads \texttt{Axis: overall}, and the
    task asks for $K$ prompts ``as different from one another as the base prompt
    allows,'' noting that \texttt{overall} names no single dimension to sweep.
    Rule 2's second sentence becomes \emph{``A prompt free to differ everywhere
    else is still not free to contradict a stated detail, and doing so is the
    single worst thing you can do here.''}
    Rule 3 is replaced by: \emph{``THE $K$ PROMPTS MUST NOT COLLAPSE, AND MUST
    NOT ALL DIFFER IN THE SAME ONE WAY. Writing the same prompt $K$ times is the
    obvious failure; $K$ prompts that differ only in, say, the time of day is
    the one to actually watch for, because it looks like an answer. Spend more
    than one degree of freedom.''} Rule 4 ends ``Add what you are changing and
    stop,'' and Rule 5 becomes: \emph{``Keep every detail the base prompt
    states, in its own wording. Everything it does not state is yours to
    change.''}
\end{itemize}

\end{tcolorbox}
\caption{Prompt used to train the generation policy with GRPO, where $K$ is the number of prompts the policy writes per completion.}
\label{fig:generation-policy-prompt}
\end{figure*}

\begin{figure*}[t]
\centering
\begin{tcolorbox}[
    colframe=black!70,
    colback=promptboxbg,
    boxrule=0.5pt
]
\footnotesize
\setlist[enumerate]{noitemsep, topsep=2pt, leftmargin=1.6em}
\setlist[itemize]{noitemsep, topsep=2pt, leftmargin=1.4em}

\textbf{Prompt Template: Editing Policy}\\[0.4em]

You are given a base prompt, the $K$ images generated from it, and a diversity
axis.

\medskip
\texttt{Base prompt: \{base\_prompt\}}\\
\texttt{Axis: \{axis\}}

\medskip
\texttt{Image 1: \textless image\textgreater}\\
\hspace*{1em}$\vdots$\\
\texttt{Image $K$: \textless image\textgreater}

\medskip
The $K$ images will be passed to an image-editing model together with one
instruction each, so that the edited images differ from one another along the
axis. Write those $K$ edit instructions.

\medskip
RULES, in order of importance:
\begin{enumerate}
    \item ONE INSTRUCTION PER IMAGE, IN ORDER. Instruction $i$ is applied to
    image $i$. Address what that image contains.

    \item EDIT, DO NOT REPLACE. Each instruction changes one aspect and leaves
    the rest of the image alone: an instruction that rewrites the whole scene
    destroys the image it was given.

    \item Vary ONLY the given axis between the $K$ instructions, and together
    cover clearly distinct values of it.

    \item THE BASE PROMPT'S CHOICES ARE CONSTRAINTS. Whatever it states
    explicitly stays true of every image after the edit: every subject, its
    count, its kind, and every attribute it names.

    \item Write each instruction as a direct command to an editing model,
    naming what to change and what it becomes: ``change the lighting to
    late-afternoon sun through a window''. Keep it under 30 words.
\end{enumerate}

\medskip
Output exactly $K$ instructions, each wrapped in
\texttt{\textless prompt\textgreater\textless /prompt\textgreater} tags, and
nothing else.

\medskip
\hrule
\medskip

\textbf{Row variants.}
\begin{itemize}
    \item \emph{\texttt{overall} rows}. The axis
    line reads \texttt{Axis: overall} and Rule 3 is replaced by: \emph{``THE $K$
    INSTRUCTIONS MUST NOT COLLAPSE, AND MUST NOT ALL CHANGE THE SAME ONE THING.
    Writing the same instruction $K$ times is the obvious failure; $K$
    instructions that all change, say, the lighting is the one to actually watch
    for, because it looks like an answer. Give each image a DIFFERENT kind of
    change --- one the setting, one the framing, one who or what is shown, one
    the light or the style --- so that the images move apart in several
    directions at once rather than further along a single one. This is still an
    edit and not a replacement: a different change per image, not a bigger one.
    Rule 2 holds.''}
\end{itemize}

\end{tcolorbox}
\caption{Prompt used to train the editing policy with GRPO where $K$ is the number of source images and edit instructions the policy writes. Unlike the generation task, the word cap is a fixed 30 words per instruction and does not depend on the renderer.}
\label{fig:edit-policy-prompt}
\end{figure*}

\clearpage
\subsection{Participant Feedback}
\label{app:user-feedback}
Table~\ref{tab:feedback} collects verbatim feedback from participants in the interactive study (Section~\ref{sec:human-eval}), grouped into overall impressions, sets they liked, and some limitations.

{\small
\setlength{\tabcolsep}{5pt}
\renewcommand{\arraystretch}{1.25}
\begin{longtable}{@{}p{0.58\linewidth}p{0.38\linewidth}@{}}
\caption{Participant feedback from the interactive studies. Quotes are verbatim, trimmed where marked. FLUX: FLUX.2 [klein]; SD3.5-M/L: SD3.5 Medium/Large; Qwen: Qwen-Image.}
\label{tab:feedback}\\
\toprule
Participant comment & What it shows / what to improve \\
\midrule
\endfirsthead
\toprule
Participant comment & What it shows / what to improve \\
\midrule
\endhead
\midrule
\multicolumn{2}{r}{\footnotesize\emph{continued on next page}} \\
\endfoot
\bottomrule
\endlastfoot

\multicolumn{2}{@{}l}{\textbf{Overall experience} (end-of-study questionnaire)} \\
\addlinespace[2pt]
``As someone who enjoys using image generators, I did purposely try to use a couple of difficult prompts that assistants often mess up, and this model did a better job than I was expecting, to be honest. Impressive.'' [FLUX] & Experienced users rate it above their expectations on hard prompts. \\
``I would say the tool overall is quite good compared to other tools I have used before for image generation.'' [SD3.5-L] & Favourable comparison with existing tools. \\
``This was really fun. I'd definitely use such a tool at a reasonable asking price.'' [Qwen] & Willingness to use, and pay for, axis-conditioned variety. \\
``[...] if I put more time in to tweaking this I could see making something like this part of my creative workflow.'' [SD3.5-L] & Fits a creative workflow. \\
``The tool covered most of the variations I needed and was easy to use.'' [Qwen] & Covers users' axes; easy to use. \\
``The task was interesting and the image variations were easy to understand.'' [Qwen] & The varied sets are legible to users. \\
``It handled everything I asked of it.'' [Qwen] & Requests are fulfilled. \\
``No it was able to realise everything i wanted.'' [SD3.5-L] & Requests are fulfilled. \\
``Really enjoyed this! I tried to mix it up with different images for different items.'' [SD3.5-L] & Engagement. \\
``I wish I could do it all over again. It was fun!'' [SD3.5-L] & Engagement; every participant asked to be invited to further studies. \\
``Thanks for the opportunity to participate in this study. I enjoyed it!'' [Qwen] & Engagement. \\

\midrule

\multicolumn{2}{@{}l}{\textbf{Sets participants liked} (comment on one set, with the participant's prompt and axis)} \\
\addlinespace[2pt]
``[...] the assistant did a great job, adhered to the prompt and provided an option I wouldn't have thought of (the stained glass one) which is easily my favorite and the one I would choose to keep.'' [FLUX] & \emph{A mechanical butterfly made of gold}; axis \emph{material}. Variety as a source of new ideas. \\
``The clothing changed clearly while the musician, guitar and city setting stayed consistent.'' [Qwen] & \emph{A street musician performing in a busy city square}; axis \emph{musician's clothing style}. The axis moves, the rest is held fixed. \\
``The variety in this is the best I've seen so far. There is life and color changes, lighting changes, and the backgrounds fit well with the people.'' [SD3.5-L] & \emph{A child running to her mother}; axis \emph{the background: city vs.\ playground vs.\ downtown vs.\ countryside}. Variety that stays coherent with the scene. \\
``Extremely impressed that it handled both the prompt and the edit requests PERFECTLY. No problems.'' [SD3.5-M] & \emph{Animated picture of a banana DJing at a rave located in a castle}; axis \emph{animation style, banana's facial expression}. Prompt and a two-part axis satisfied together. \\
``Really impressed for how well the image generator handled all of the changes I requested.'' [Qwen] & \emph{A preacher with face tattoos playing video games inside of a cave}; axis \emph{cave, angle, controller, face tattoos, ethnicity, facial expression}. Handles a six-part request. \\
\midrule

\multicolumn{2}{@{}l}{\textbf{Limitation: viewpoint, framing and distance}} \\
\addlinespace[2pt]
``It would not change the angle of the pictures to any significant extent. I tried editing these, generating new, changing the prompt too, and it would not seem to work.'' [SD3.5-M] & \multirow{2}{=}{Camera axes (angle, framing, zoom) are the most frequent failure. Train on viewpoint-specific rows and reward view change in image space; check how far each renderer honours view terms.} \\
``The cookies varied in type and style, the kitchens varied [...]. The only issue was all of the images kept the same basic angle/camera position.'' [SD3.5-L] & \\
\midrule

\multicolumn{2}{@{}l}{\textbf{Limitation: preserving the prompt at high variety}} \\
\addlinespace[2pt]
``[...] the variability is much better, but the pictures lost my core instruction that the alien be ON a cloud and the people NOT know she is there.'' [SD3.5-L] & \multirow{2}{=}{Stronger variety drops stated constraints or adds content. Check each constraint of the base prompt, not only set-level alignment, and penalise unrequested changes.} \\
``It changed woman, expressions and colors of the clothing, but it also changed hair color which I didn't ask and was a crucial part of the prompt.'' [Qwen] & \\
\midrule

\multicolumn{2}{@{}l}{\textbf{Limitation: spread along the requested axis}} \\
\addlinespace[2pt]
``They varied the right thing, but all the styles of Yukatas were only varied with colors but still very similar in style [...].'' [Qwen] & \multirow{2}{=}{The axis moves, but narrowly or with repeats. Reward coverage of distinct values along the axis (e.g.\ four different ethnicities), not any change.} \\
``They correctly varied the races, but it's interesting that two women are Asian instead of providing a greater variety.'' [SD3.5-L] & \\
\midrule

\multicolumn{2}{@{}l}{\textbf{Limitation: style and medium}} \\
\addlinespace[2pt]
``Instead of changing the illustration style, the ai has changed the architectural features of the original drawings which is not the variation I was seeking.'' [SD3.5-M] & \multirow{2}{=}{Medium changes (watercolour, sketch, animation style) are often realised as content changes. Place style terms prominently and measure style separately from content.} \\
``I wanted to see different artistic styles, not just different outfits.'' [SD3.5-L] & \\
\midrule

\multicolumn{2}{@{}l}{\textbf{Limitation: pose and action}} \\
\addlinespace[2pt]
``I wish the images were less static and more dynamically oriented. In all four images, the pair is simply facing the camera with no large emphasis on action.'' [Qwen] & \multirow{2}{=}{Pose and action are under-varied. Add pose and action axes to the training mix.} \\
``[...] the poses didn't change, I expected there to be maybe some wings flapping or even a bird in flight.'' [SD3.5-L] & \\
\midrule

\multicolumn{2}{@{}l}{\textbf{Limitation: implicit intent and context}} \\
\addlinespace[2pt]
``[...] it appears to not pick up the implicit intent as well as it does with the explicit intent of the user.'' [Qwen] & \multirow{2}{=}{Infer what a request implies (e.g.\ varying animals implies varying their size and pose), and read context in the prompt.} \\
``I would have expected the poses of varied animals to be more different to each other, and the sizes of the animals to be more varied overall. I know this isn't directly specified in the axis, but feel like its implied [...].'' [Qwen] & \\
\midrule

\multicolumn{2}{@{}l}{\textbf{Limitation: what the renderer can express}} \\
\addlinespace[2pt]
``I wanted it to show numbers 1-24 instead of 1-12, but it only showed clocks with 12 numbers.'' [SD3.5-L] & \multirow{2}{=}{Counts, rare concepts and specific products fail already in i.i.d.\ samples. Detect when the renderer cannot express a request and flag it rather than vary around it.} \\
``I could not get the model to only place 4 female dancers and 1 male dancer on the stage. There was always at least one photo with 3 or 5 female dancers.'' [Qwen] & \\
\midrule

\multicolumn{2}{@{}l}{\textbf{Limitation: image quality under repeated editing}} \\
\addlinespace[2pt]
``The pictures changed according to the axes, but the artifacts that were already there increased, making them unusable in practice.'' [SD3.5-M] & \multirow{2}{=}{Artifacts accumulate over edit rounds of an image-to-image editor. Limit edit depth, or re-render from the rewritten prompt after several edits.} \\
``I tried "Middle," "More," and "Most," but the formation shape didn't change the way I wanted, and artifacts gradually increased.'' [SD3.5-M] & \\
\midrule

\multicolumn{2}{@{}l}{\textbf{Limitation: varying from one chosen image}} \\
\addlinespace[2pt]
``The main thing that I would want the vary tool to do here, is take one image as the base and vary from that.'' [FLUX] & \multirow{2}{=}{The most requested feature. Condition the policy and editor on one selected image and return four variations of it along the axis.} \\
``I would have liked to have been able to vary one picture at a time rather than having the tool decide which photos to vary [...].'' [SD3.5-L] & \\
\midrule

\multicolumn{2}{@{}l}{\textbf{Limitation: keeping parts of an image}} \\
\addlinespace[2pt]
``There were certain aspects of generated images I would like to keep, like the specific position of a person or object but I could only keep the entire image or nothing at all.'' [Qwen] & \multirow{2}{=}{Region-level control: keep or change parts of an image, not only whole images.} \\
``One of the images produced cartoon animals and I wanted to keep this particular image but delete the cartoon animals but was unable to.'' [Qwen] & \\
\midrule

\multicolumn{2}{@{}l}{\textbf{Limitation: controlling the strength of variety}} \\
\addlinespace[2pt]
``The slider function seems to be either very subtle and fine or outrageous.'' [FLUX] & \multirow{2}{=}{Make strength predictable: calibrate the scale per renderer and allow a separate strength per attribute.} \\
``I did want to vary multiple factors in an image to different degrees (i.e.\ high variance in the chef's race/gender, but low variance in the outfit), and I couldn't control the variance of the different factors.'' [SD3.5-L] & \\
\end{longtable}
}

\begin{figure*}[p]
\centering
\includegraphics[width=\textwidth]{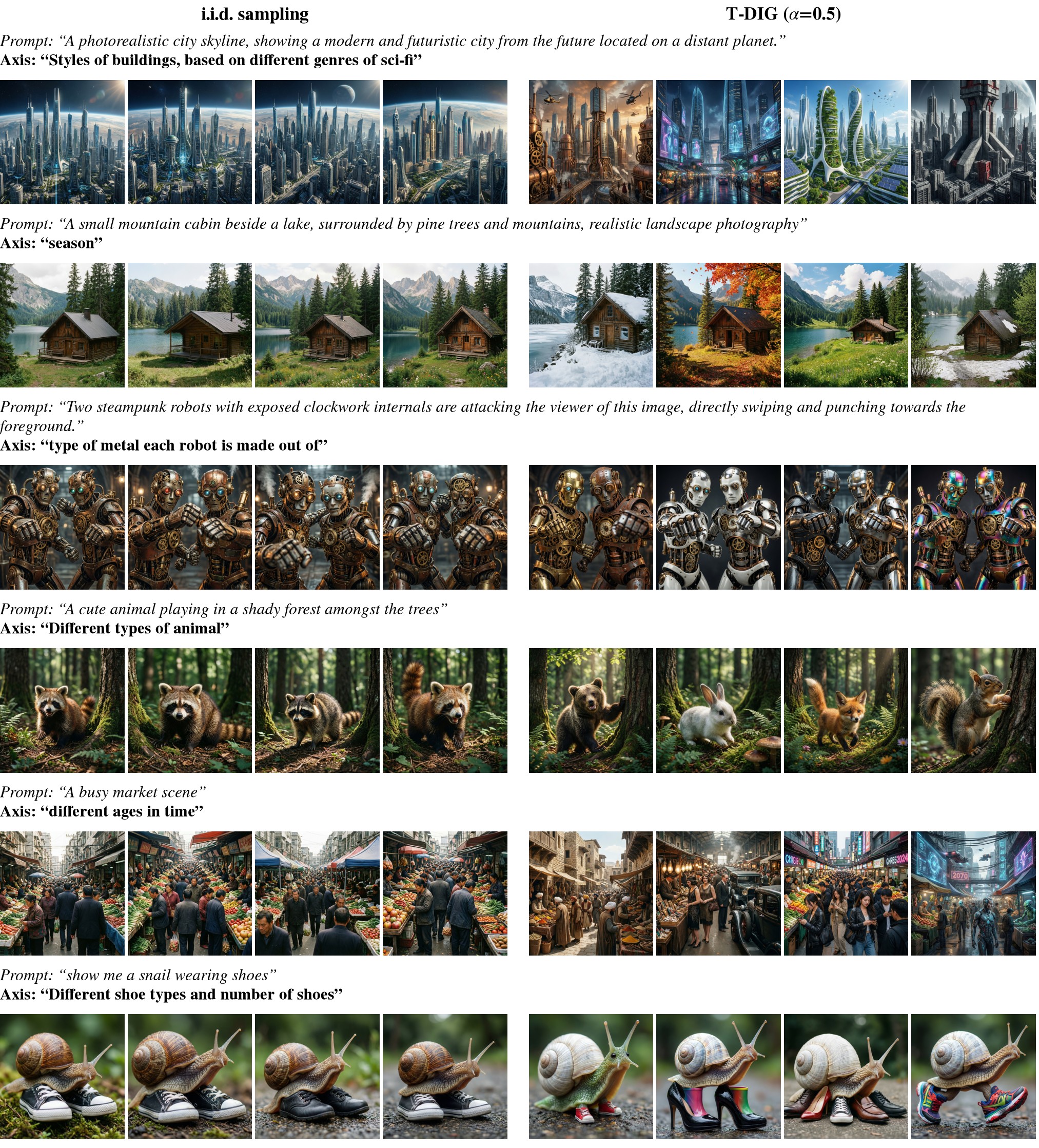}
\caption{Qwen-Image (1/3): i.i.d.\ sampling (left) vs.\ T-DIG ($\alpha$=0.5, right) on requests from the interactive test set. Prompts and axes are the study participants' own text, verbatim.}
\label{fig:gallery-qwen-1}
\end{figure*}

\begin{figure*}[p]
\centering
\includegraphics[width=\textwidth]{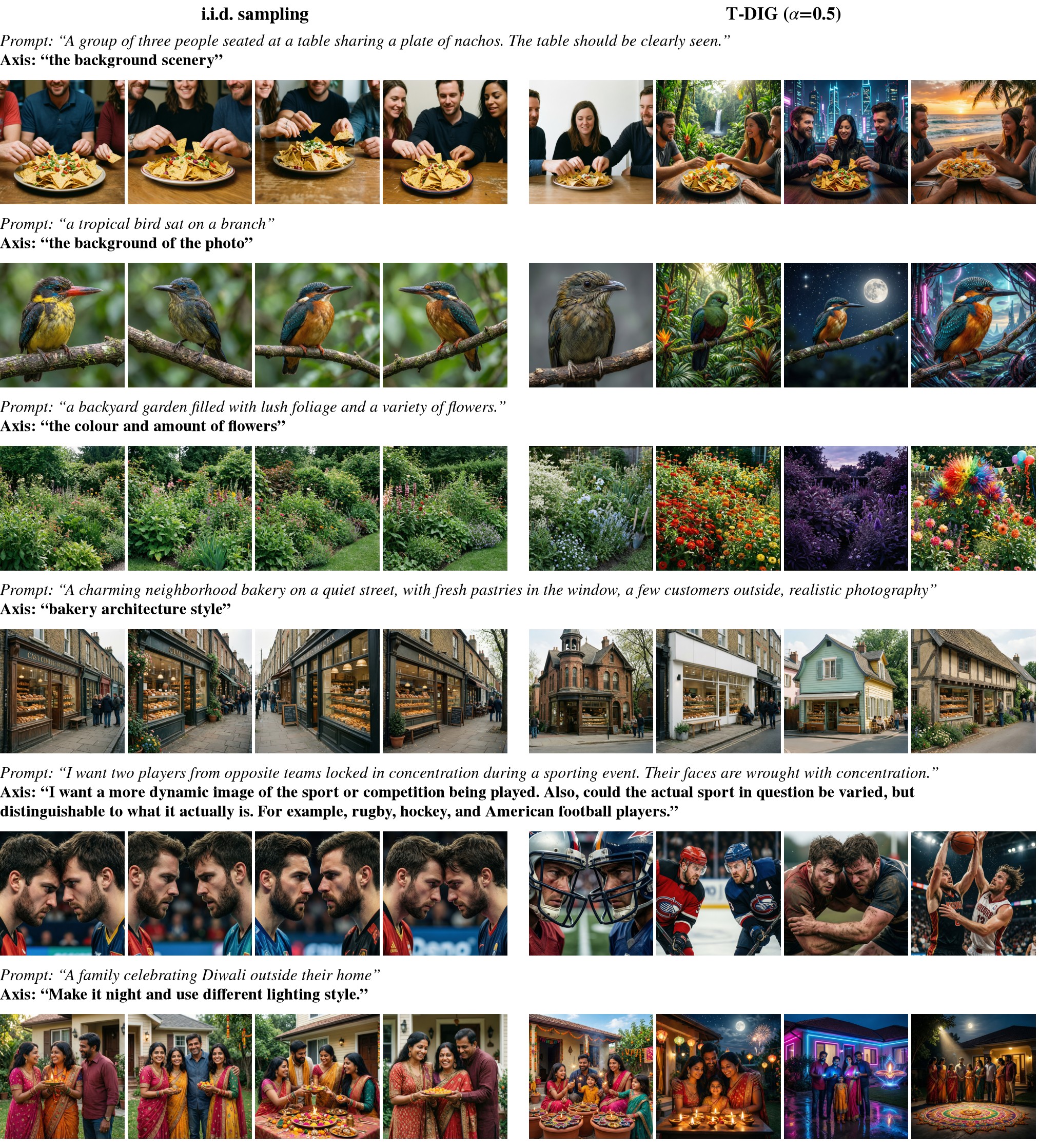}
\caption{Qwen-Image (2/3): i.i.d.\ sampling (left) vs.\ T-DIG ($\alpha$=0.5, right) on requests from the interactive test set. Prompts and axes are the study participants' own text, verbatim.}
\label{fig:gallery-qwen-2}
\end{figure*}

\begin{figure*}[p]
\centering
\includegraphics[width=\textwidth]{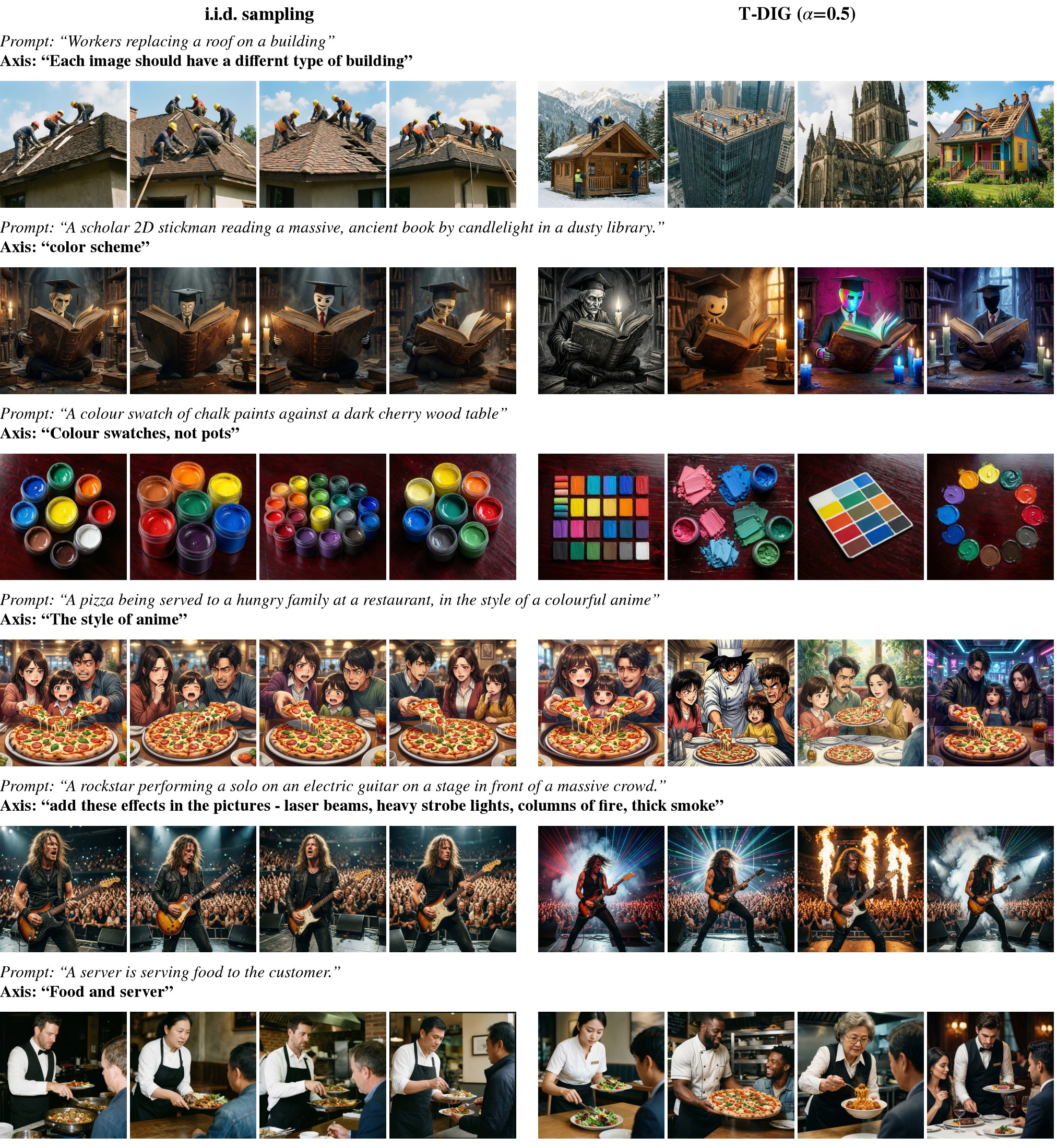}
\caption{Qwen-Image (3/3): i.i.d.\ sampling (left) vs.\ T-DIG ($\alpha$=0.5, right) on requests from the interactive test set. Prompts and axes are the study participants' own text, verbatim.}
\label{fig:gallery-qwen-3}
\end{figure*}

\begin{figure*}[p]
\centering
\includegraphics[width=\textwidth]{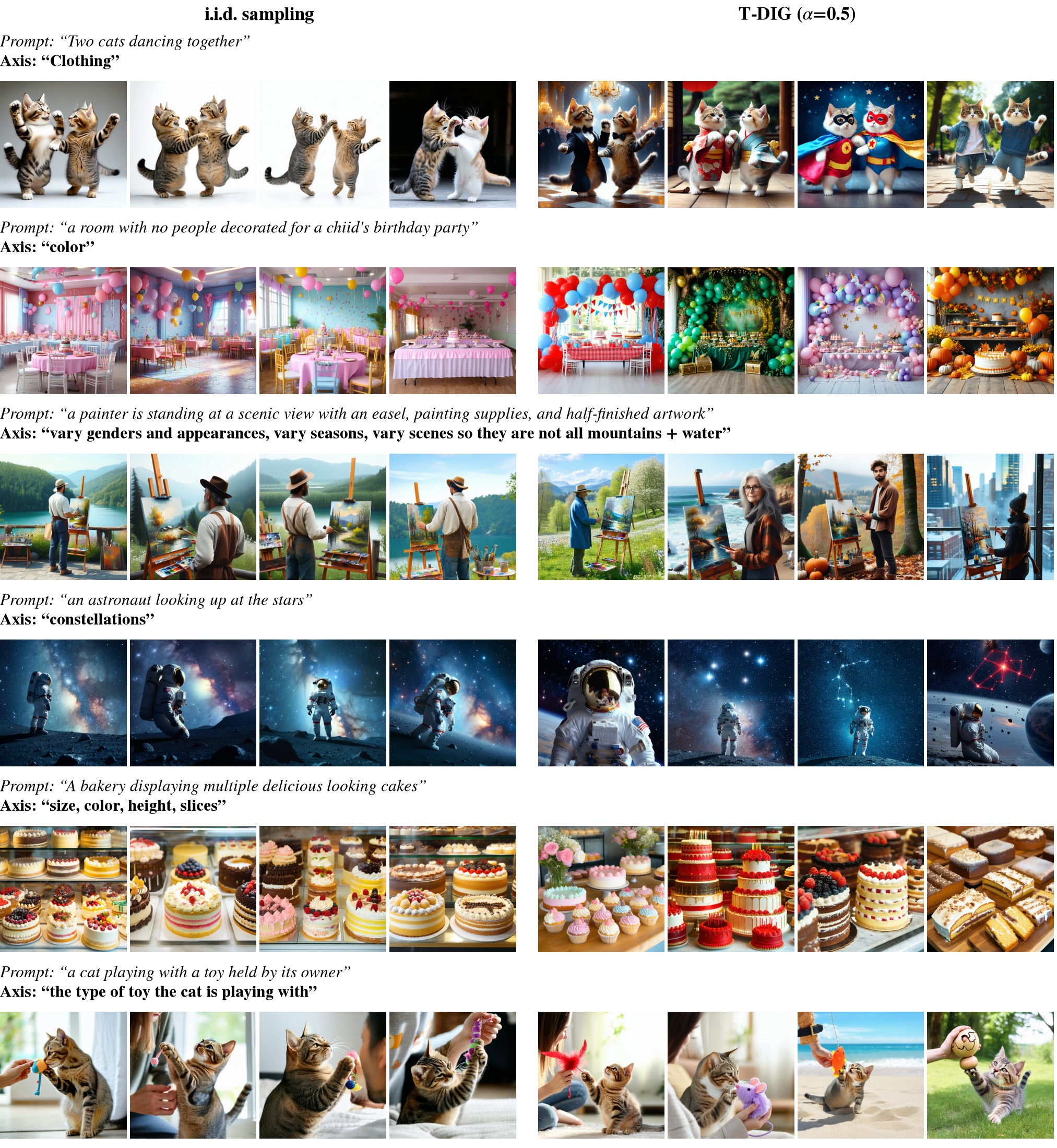}
\caption{SD3.5 L (1/3): i.i.d.\ sampling (left) vs.\ T-DIG ($\alpha$=0.5, right) on requests from the interactive test set. Prompts and axes are the study participants' own text, verbatim.}
\label{fig:gallery-sdl-1}
\end{figure*}

\begin{figure*}[p]
\centering
\includegraphics[width=\textwidth]{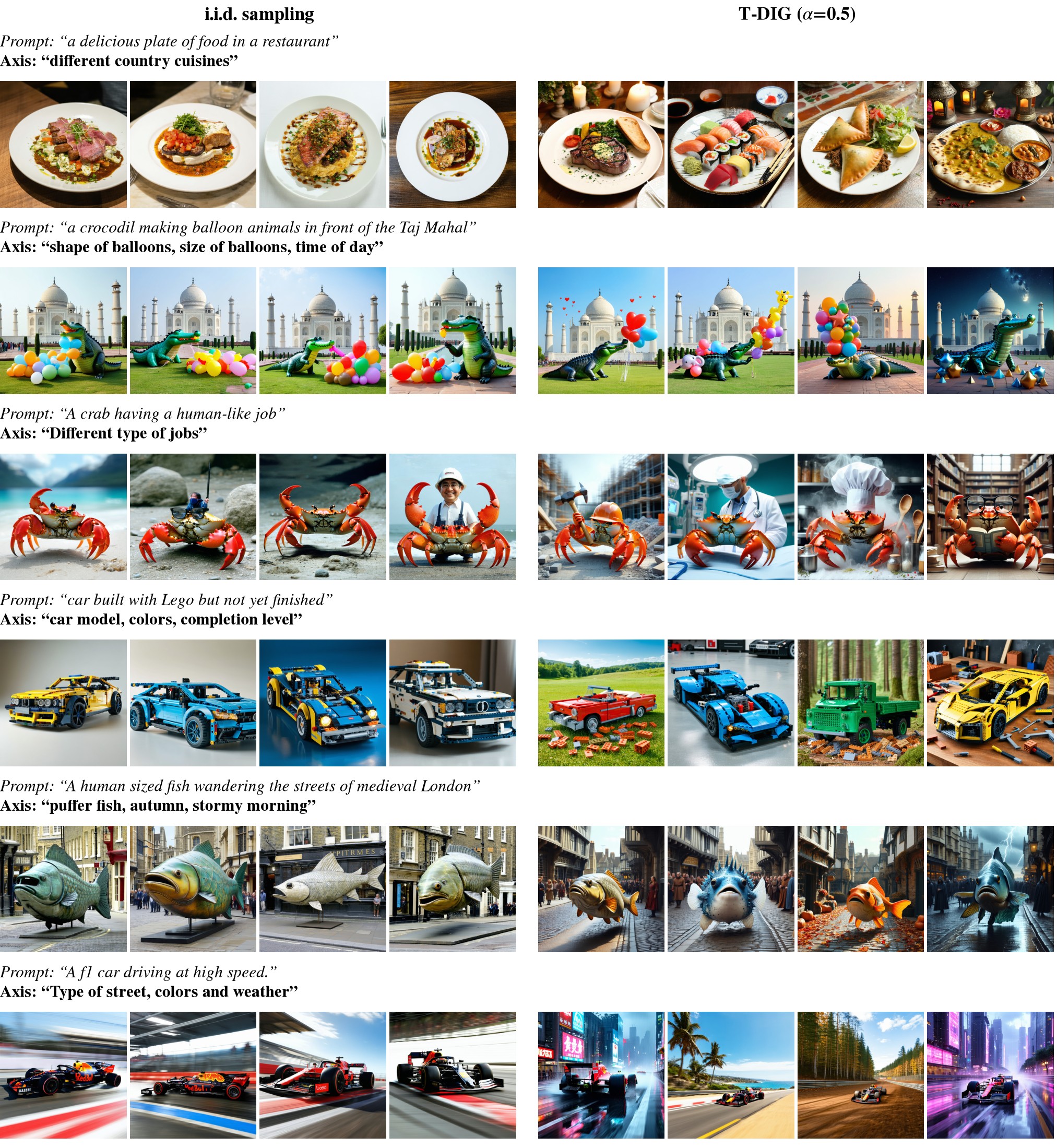}
\caption{SD3.5 L (2/3): i.i.d.\ sampling (left) vs.\ T-DIG ($\alpha$=0.5, right) on requests from the interactive test set. Prompts and axes are the study participants' own text, verbatim.}
\label{fig:gallery-sdl-2}
\end{figure*}

\begin{figure*}[p]
\centering
\includegraphics[width=\textwidth]{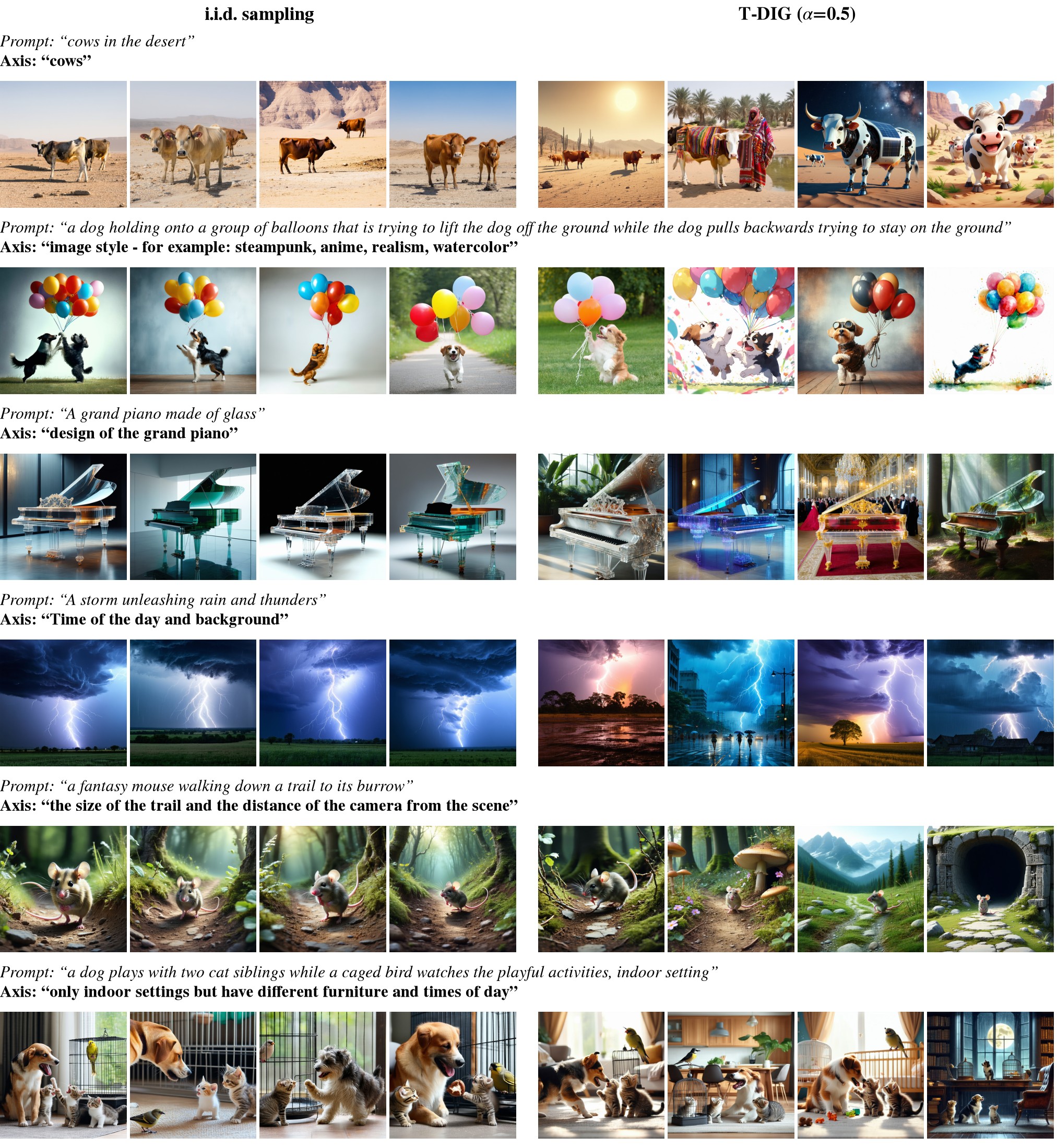}
\caption{SD3.5 L (3/3): i.i.d.\ sampling (left) vs.\ T-DIG ($\alpha$=0.5, right) on requests from the interactive test set. Prompts and axes are the study participants' own text, verbatim.}
\label{fig:gallery-sdl-3}
\end{figure*}

\begin{figure*}[p]
\centering
\includegraphics[width=\textwidth]{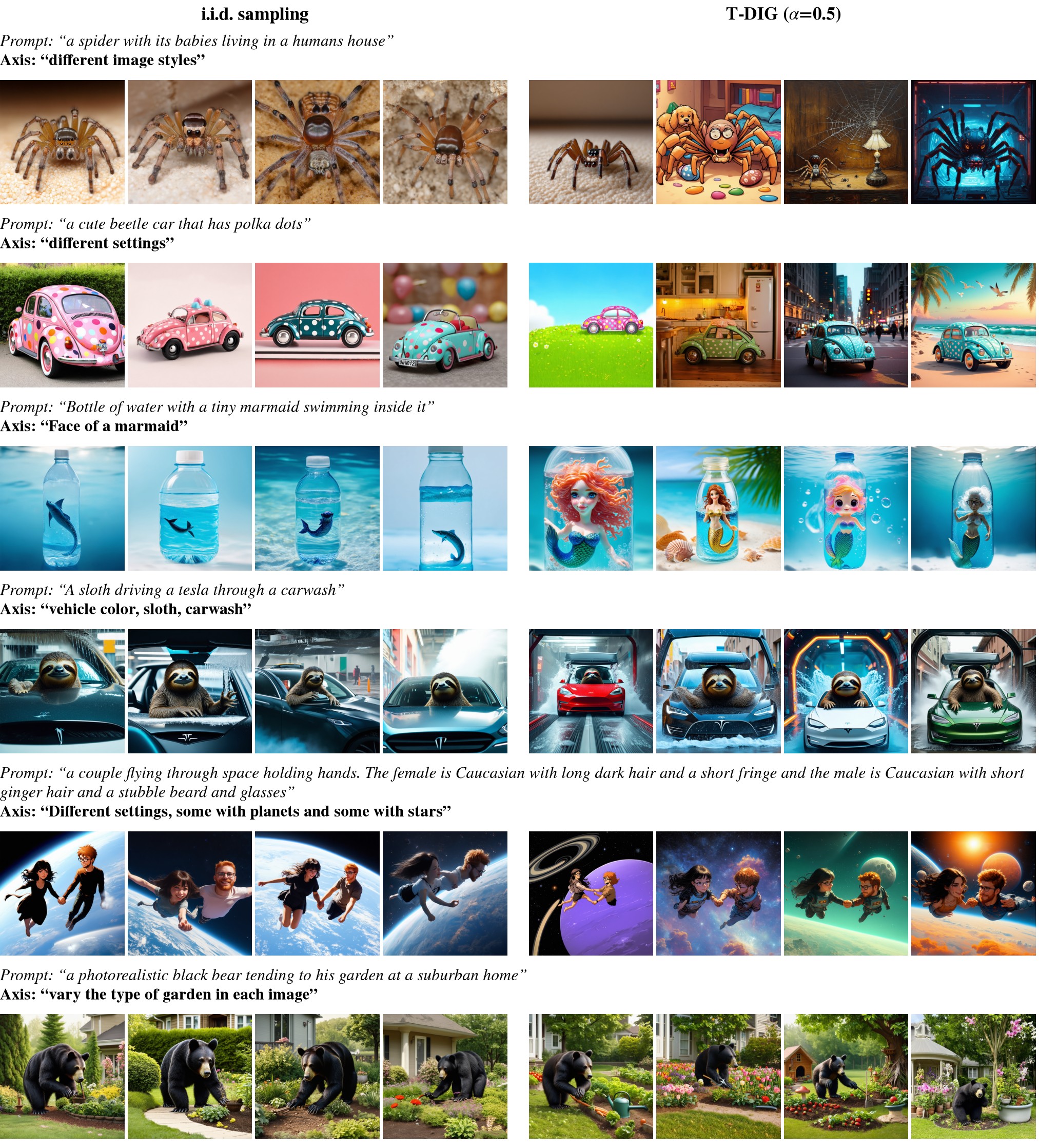}
\caption{SD3.5 M (1/2): i.i.d.\ sampling (left) vs.\ T-DIG ($\alpha$=0.5, right) on requests from the interactive test set. Prompts and axes are the study participants' own text, verbatim.}
\label{fig:gallery-sdm-1}
\end{figure*}

\begin{figure*}[p]
\centering
\includegraphics[width=\textwidth]{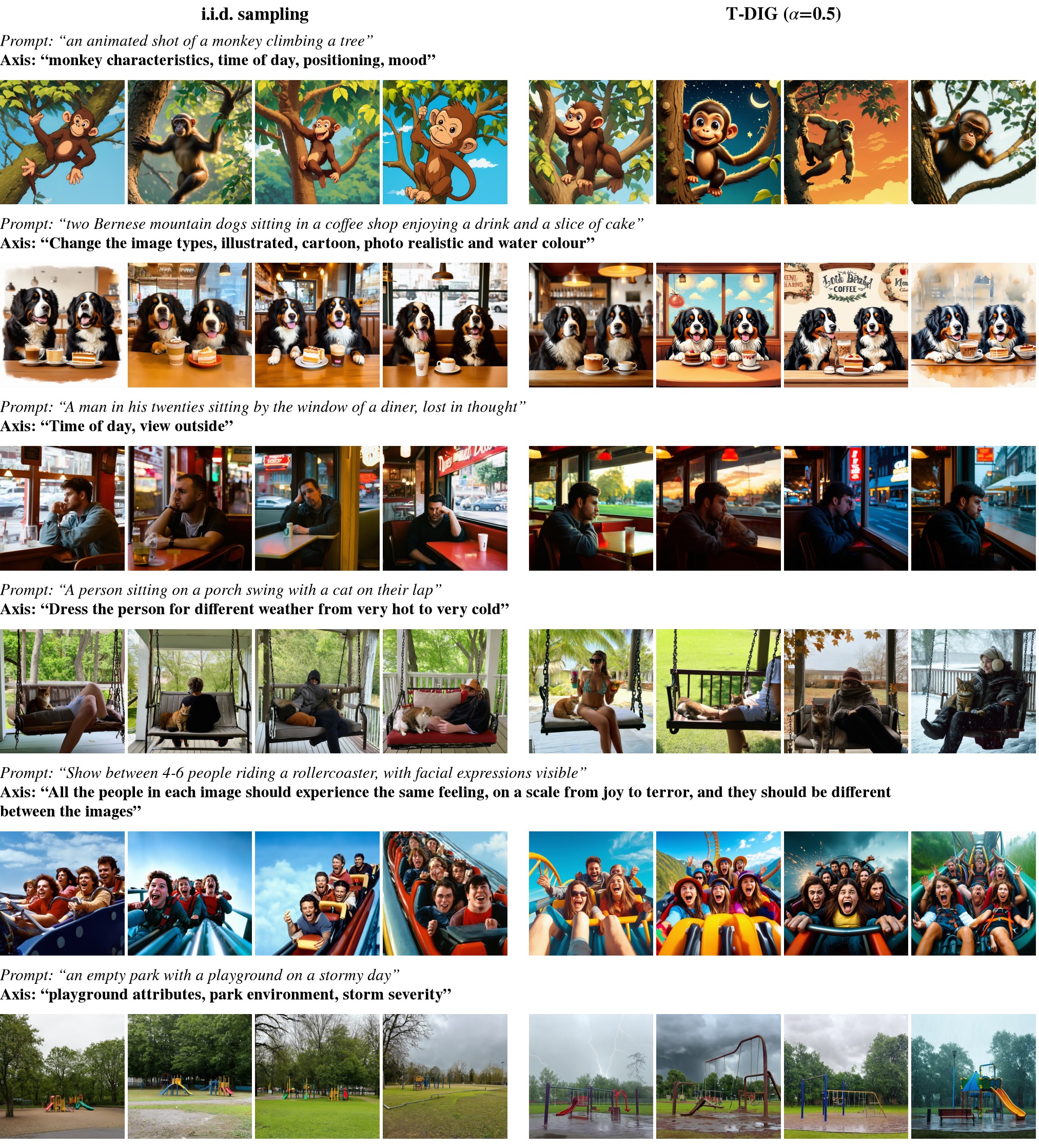}
\caption{SD3.5 M (2/2): i.i.d.\ sampling (left) vs.\ T-DIG ($\alpha$=0.5, right) on requests from the interactive test set. Prompts and axes are the study participants' own text, verbatim.}
\label{fig:gallery-sdm-2}
\end{figure*}

\begin{figure*}[p]
\centering
\includegraphics[width=\textwidth]{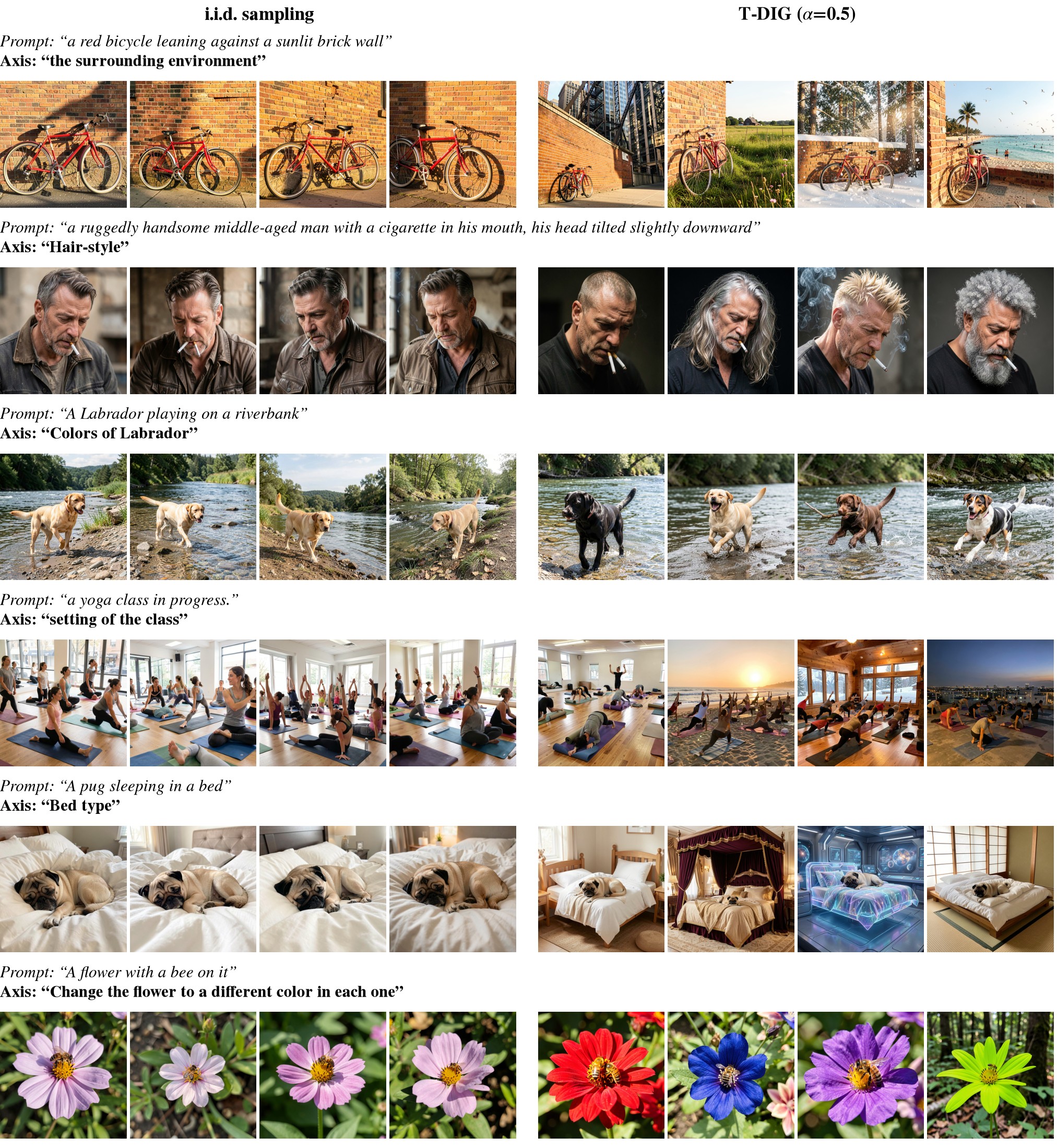}
\caption{FLUX.2 [klein] (1/3): i.i.d.\ sampling (left) vs.\ T-DIG ($\alpha$=0.5, right) on requests from the interactive test set. Prompts and axes are the study participants' own text, verbatim.}
\label{fig:gallery-flux-1}
\end{figure*}

\begin{figure*}[p]
\centering
\includegraphics[width=\textwidth]{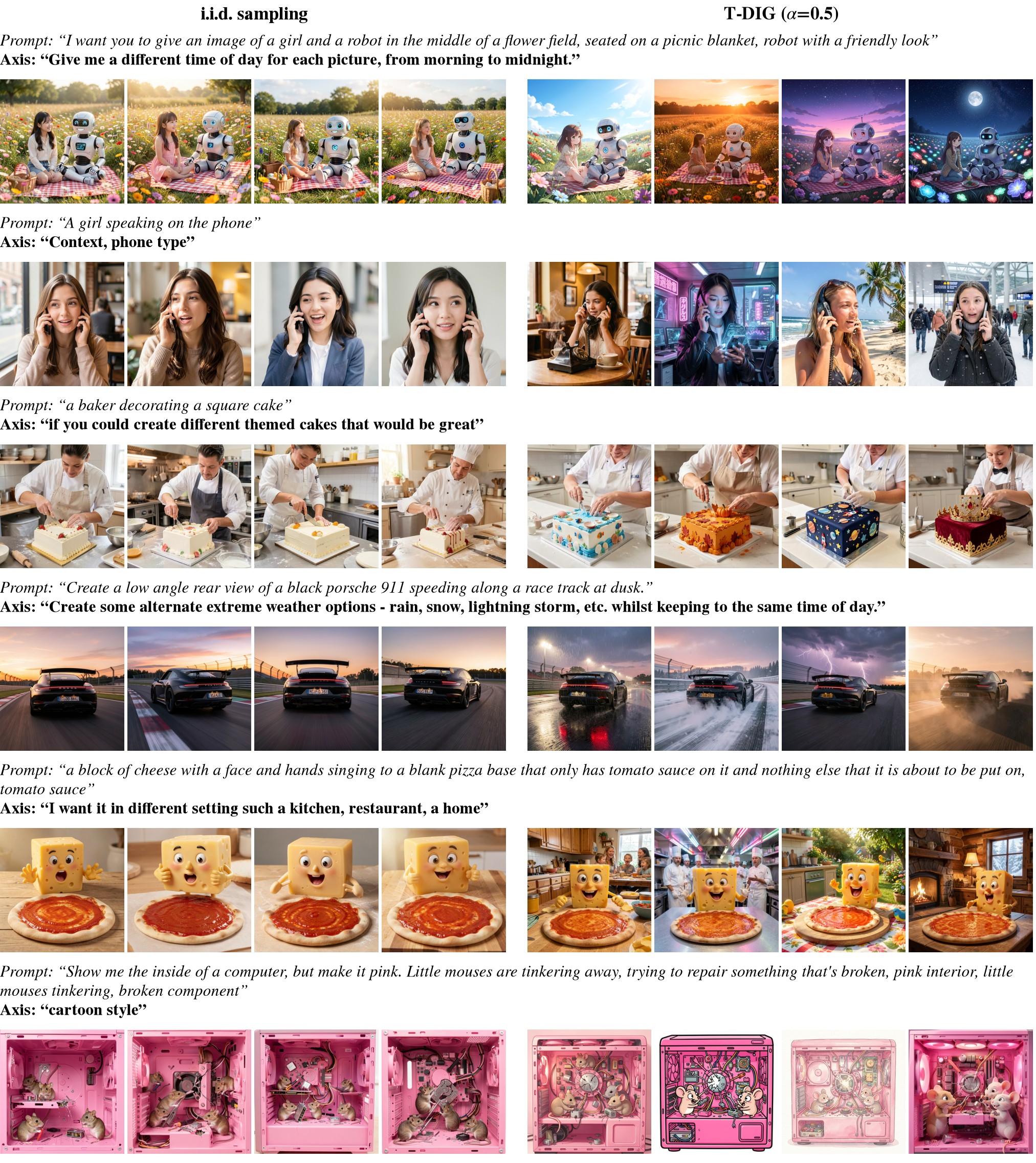}
\caption{FLUX.2 [klein] (2/3): i.i.d.\ sampling (left) vs.\ T-DIG ($\alpha$=0.5, right) on requests from the interactive test set. Prompts and axes are the study participants' own text, verbatim.}
\label{fig:gallery-flux-2}
\end{figure*}

\begin{figure*}[p]
\centering
\includegraphics[width=\textwidth]{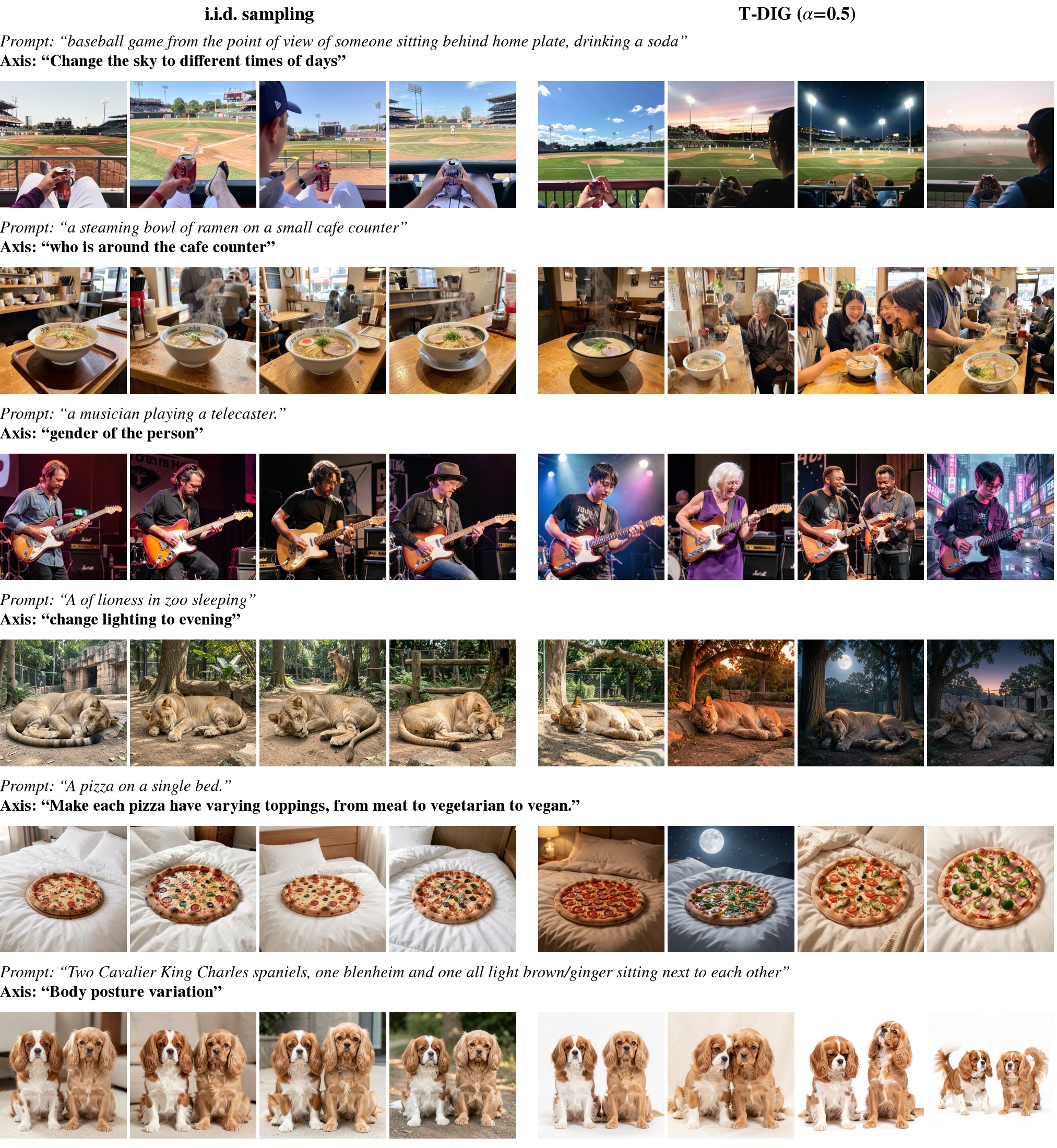}
\caption{FLUX.2 [klein] (3/3): i.i.d.\ sampling (left) vs.\ T-DIG ($\alpha$=0.5, right) on requests from the interactive test set. Prompts and axes are the study participants' own text, verbatim.}
\label{fig:gallery-flux-3}
\end{figure*}
